\documentclass[11pt]{article}

\usepackage[preprint]{acl}

\usepackage{times}
\usepackage{latexsym}
\usepackage[T1]{fontenc}
\usepackage[utf8]{inputenc}
\usepackage{microtype}
\usepackage{inconsolata}
\usepackage{graphicx}
\usepackage{amsmath}
\usepackage{tabularx}
\usepackage{booktabs}
\usepackage{multirow}
\usepackage{tabularx}
\usepackage{float}
\usepackage{amsfonts}
\usepackage{comment}
\usepackage{xcolor}
\usepackage{framed}
\usepackage[most]{tcolorbox}
\usepackage{soul}
\sethlcolor{yellow}
\title{Temporal Flattening in LLM-Generated Text: Comparing Human and LLM Writing Trajectories}

\author{
  Zhanwei Cao\thanks{Authors are listed alphabetically.}
              \thanks{The student authors contributed equally.} \and
  Yeojin Go\footnotemark[1]\footnotemark[2] \and
  Yifan Hu\footnotemark[1]\thanks{The senior authors contributed equally.} \and
  Shanu Sushmita\footnotemark[1]\footnotemark[3] \\
  Northeastern University \\
  Seattle, WA, USA \\
  \texttt{\{cao.zha, kim.yeojin,  yif.hu, s.sushmita\}@northeastern.edu}
}

\begin{document}
\maketitle

% ================= Abstract =================

\begin{abstract}
Large language models (LLMs) are increasingly used in daily applications, from content generation to code writing, where each interaction treats the model as stateless, generating responses independently without memory. Yet human writing is inherently longitudinal: authors' styles and cognitive states evolve across months and years. This raises a central question: can LLMs reproduce such temporal structure across extended time periods? We construct and publicly release a longitudinal dataset of 412 human authors and 6,086 documents spanning 2012--2024 across three domains (academic abstracts, blogs, news) and compare them to trajectories generated by three representative LLMs under standard and history-conditioned generation settings. Using drift and variance-based metrics over semantic, lexical, and cognitive--emotional representations, we find \emph{temporal flattening} in LLM-generated text. LLMs produce greater lexical diversity but exhibit substantially reduced semantic and cognitive--emotional drift relative to humans. These differences are highly predictive: temporal variability patterns alone achieve 94\% accuracy and 98\% ROC-AUC in distinguishing human from LLM trajectories. Our results demonstrate that temporal flattening persists regardless of whether LLMs generate independently or with access to incremental history, revealing a fundamental property of current deployment paradigms. This gap has direct implications for applications requiring authentic temporal structure, such as synthetic training data and longitudinal text modeling\footnote{Dataset available at \url{https://github.com/yjkim717/Cognitive-Emotional-Trajectories}}.
\end{abstract}

% ================= Main Sections =================
\section{Introduction}

Large Language Models (LLMs) are now capable of generating text that is fluent, coherent, and stylistically controlled 
\cite{NEURIPS2020_1457c0d6}. Consequently, they are widely deployed across diverse domains, including content 
generation, dialogue systems, and educational tutoring 
\cite{NEURIPS2022_b1efde53, touvron2023llama2openfoundation}. Of particular relevance, LLMs are increasingly used for synthetic training data production \cite{bai2022constitutionalaiharmlessnessai}. In these applications, LLMs generate outputs independently for each query, often conditioned on attributes like topic, length, or style, treating each generation as a standalone instance.

However, human writing is inherently longitudinal. What a person wrote previously and how their writing has evolved over time carries significant meaning, reflecting evolving perspectives, shifting emphases, and changing life circumstances \cite{miaschi-etal-2020-tracking, 10.1162/coli_a_00428, tsakalidis-etal-2022-identifying}. Prior psycholinguistic and computational studies demonstrate that 
stylistic habits, emotional tone, and linguistic complexity drift 
across months or years \cite{miaschi-etal-2020-tracking, 10.1162/coli_a_00428, tsakalidis-etal-2022-identifying, 
Tausczik2010ThePM}, forming a temporal structure that is intrinsic to human communication. Yet current LLM deployment practices do not account for this longitudinal structure.

Standard approaches for LLM-based text generation rely on instance-wise 
sampling \cite{NEURIPS2022_b1efde53, patel-etal-2024-datadreamer} under the implicit 
assumption that independently generated documents adequately approximate 
the distribution of human writing. However, this paradigm risks collapsing the longitudinal structure inherent in authentic authorship—properties essential to tasks requiring temporal coherence. In this work, we investigate two fundamental research questions:

\textbf{RQ1:} When LLMs generate documents across extended time periods, can they reproduce human-like temporal structure?

\textbf{RQ2:} If there is a difference, what aspects of human writing dynamics are systematically lost or flattened?

To answer these questions, we analyze human authors' writing trajectories across three domains (academic abstracts, blogs, news) and compare them to trajectories generated by independently sampling from LLMs. For each document in a human author's timeline, we prompt an LLM with specifications matched on domain, topic, and length, either without prior context (instance-wise) or with cumulative summaries of previous outputs (history-augmented).

Our methodology investigates whether LLMs can reproduce the temporal dynamics inherent in human writing across multiple dimensions. We measure evolution in three aspects: cognitive--emotional features (personality, sentiment, style), lexical representations, and semantic embeddings. For each dimension, we quantify temporal structure through two complementary metrics: drift (cumulative shifts over time) and variance (fluctuation irregularity across the trajectory). We assess whether humans exhibit significantly greater evolution than LLMs using matched-pair binomial tests, and identify which cognitive--emotional features most distinguish human temporal patterns through predictive probes with machine learning classifiers.

Our contributions are as follows:
\begin{itemize}
    \item We construct and publicly release a longitudinal dataset of 412 human 
    authors and 6,086 documents spanning 2012--2024 across three domains 
    (academic abstracts, blogs, news).
    \item We identify systematic \emph{temporal flattening} in LLM-generated 
    text: LLMs produce high lexical diversity but substantially reduced semantic 
    and cognitive--emotional drift, with temporal variability patterns alone 
    achieving 94\% classification accuracy.
    \item We determine which cognitive--emotional features (personality traits, 
    stylistic complexity, sentiment) most distinguish human from LLM temporal 
    dynamics.
    \item We demonstrate that temporal flattening persists under both instance-wise and history-augmented generation, revealing a systematic property of current deployment practices rather than a prompt-design artifact, with direct implications for applications requiring authentic temporal dynamics such as synthetic training data and longitudinal text modeling.
\end{itemize}
\section{Related Work}

\textbf{Temporal dynamics in human writing.} Research on temporal text analysis has documented that human writing exhibits measurable evolution over time. Miaschi et al.~\cite{miaschi-etal-2020-tracking} showed syntactic complexity growth in second-language learners, while Vincze et al.~\cite{10.1162/coli_a_00428} demonstrated that linguistic features reflect cognitive--emotional decline in patients with Alzheimer's disease. Tsakalidis et al.~\cite{tsakalidis-etal-2022-identifying} proposed longitudinal evaluation methods to detect ``moments of change'' in user text, capturing shifts in affective expression across individual timelines. Diachronic analyses of literature \cite{rios-toledo2022detection} and journalism \cite{mccarthy-dore-2022-hong} further confirm that stylistic patterns evolve across decades. Computational stylometry research has similarly demonstrated that writing style shifts are detectable over multi-year intervals \cite{Can2004ChangeOW}.

\textbf{Human--LLM co-evolution.} Recent evidence suggests human writing is already adapting to LLM influence. Geng and Trotta~\cite{geng-trotta-2025-human} analyzed arXiv abstracts and found that words strongly associated with ChatGPT (e.g., ``delve,'' ``intricate'') decreased sharply once flagged by researchers, while common academic terms continued rising. Kousha \& Thelwall~\cite{kousha2025llminfluence} and Liang et al.~\cite{liang2024llmusage} documented similar patterns across scientific literature. However, these analyses remain largely lexical and do not examine broader cognitive--emotional or semantic dynamics.

\textbf{Synthetic data generation.}
LLMs are increasingly used to produce training data for fine-tuning \cite{long-etal-2024-llms}, data augmentation \cite{patel-etal-2024-datadreamer}, and benchmark construction \cite{li-etal-2023-synthetic}.
Most work relies on \emph{instance-wise} synthesis, generating each sample independently from attributes such as topic, length, or style \cite{chim-etal-2025-evaluating}.
While this paradigm performs well on static quality metrics (e.g., Chim et al.\ show semantic content is largely preserved), it has systematic limitations in maintaining stylistic fidelity and fails to model temporal coherence.

This limitation has critical implications for downstream applications requiring longitudinal understanding. Models trained on temporally incoherent synthetic data may fail to capture stylometric evolution in authorship attribution \cite{huang-etal-2024-large}, replicate cognitive--emotional shifts in mental health trajectory modeling, or track preference drift in adaptive personalization systems. Recent work on model collapse \cite{shumailov2024collapse,Gerstgrasser2024IsMC} and synthetic data quality \cite{kang-etal-2025-demystifying} further underscores these risks: models trained on such data may inherit not only static distributional biases but also fundamental misrepresentations of how language and cognition evolve over time.

\textbf{Temporal representation learning.}
Methods such as dynamic word embeddings \cite{hamilton-etal-2016-diachronic}, diachronic transformers, and trajectory-based frameworks \cite{10.1145/3656470} model how language evolves over time and offer tools for quantifying semantic drift. Building on this foundation, we ask whether LLMs—when conditioned only on topic and length—can reproduce the cognitive, emotional, and stylistic dynamics present in human writing, and whether systematic differences expose optimization asymmetries between surface lexical diversity and deeper semantic evolution.

\textbf{Temporal modeling in large language models.}
Recent studies have begun examining how LLMs interact with temporal 
information in text. Zhu et al.~\cite{chenghaozhu-etal-2025-llm} 
show that LLM performance often degrades when evaluated on time-stamped 
benchmarks from unseen periods. Wu and Pan~\cite{wu2025dynamictopicevolutiontemporal} 
investigate how topical patterns shift over time, while Wang et 
al.~\cite{wang2025timefeatureexploitingtemporal} highlight the importance of temporal 
signals when generating or evaluating chronologically grounded text. 
Related work has also explored temporal dependencies in event sequences 
\cite{kong2026bytetokenenhancedlanguagemodels} and temporal structure in text-graph settings 
\cite{zhang2025unifyingtextsemanticsgraph}.

These studies primarily analyze temporal behavior at the corpus or 
event level. In contrast, our work focuses on \emph{per-author} 
longitudinal evolution and evaluates whether LLM 
generation, both instance-wise and with access to incremental history, can reproduce the 
year-to-year semantic and cognitive–-emotional dynamics characteristic 
of human writing.
\section{Dataset}
\label{sec:dataset}
We constructed and curated a longitudinal corpus of 412 human authors spanning
2012--2024 across three domains: academic abstracts, blogs, and news. After preprocessing, the corpus contains 6,086 human-authored documents. For each human author, we generated
matched trajectories from three commercial LLMs—DeepSeek V1, GPT--4o mini
(\texttt{gpt-4o-mini-2024-07-18}), and Claude 3.5 Haiku
(\texttt{claude-3-5-haiku-20241022})—yielding 103,459 LLM-generated documents across both instance-wise and history-augmented conditions. This dataset, including human texts, synthetic outputs, metadata, and
prompts, is publicly available\footnote{\url{https://github.com/yjkim717/Cognitive-Emotional-Trajectories}} to support reproducibility.

\paragraph{Human data.}
Human documents were collected from three domains, each requiring authors to 
have at least five consecutive years of writing:
\begin{itemize}
    \item \textbf{Academic abstracts (2020--2024):} 100 authors from five STEM fields, 
    500 documents total, collected via ACL Anthology, PubMed, and Crossref APIs. See Appendix~\ref{app:acq-academic}
    \item \textbf{Blogs (2020--2024):} 195 Reddit users across four topical domains 
    (Social, Lifestyle, Sports, Technology), 1,901 posts total, collected via PRAW. See Appendix~\ref{app:acq-reddit}
    \item \textbf{News (2012--2022):} 117 journalists from the HuffPost News Category 
    Dataset, 3,685 articles total. See Appendix~\ref{app:acq-news}
\end{itemize}

\paragraph{LLM-generated data.}
For each human author, we generate matched trajectories using three prompting configurations of increasing specificity: (1) minimal prompts conditioned only on keywords, topic, and target length; (2) prompts that additionally specify a genre-appropriate persona (e.g., researcher, journalist, blogger); and (3) prompts that combine persona with 1--2 human examples for style guidance. Each LLM trajectory is strictly matched to its corresponding human author: for every human document, we generate one LLM document conditioned on the same domain, extracted topic keywords, and target length, ensuring identical yearly document counts and topical alignment. In the instance-wise condition, all configurations exclude prior outputs and writing history, mirroring standard synthetic data generation practices. In the history-augmented condition, cumulative yearly summaries of prior LLM outputs are prepended to the prompt (Appendix~\ref{app:prompts-history}). To test robustness across model families and prompting strategies, multiple synthetic trajectories are generated per author. Full prompting templates and generation protocols appear in Appendix~\ref{app:prompts}.

Figure~\ref{fig:trajectory-generation} illustrates the key distinction: human trajectories reflect continuous authorship informed by writing history, whereas LLM trajectories are generated either independently (instance-wise) or with access to cumulative summaries of prior outputs (history-augmented).

\begin{figure}[t]
\centering
\includegraphics[width=\columnwidth]{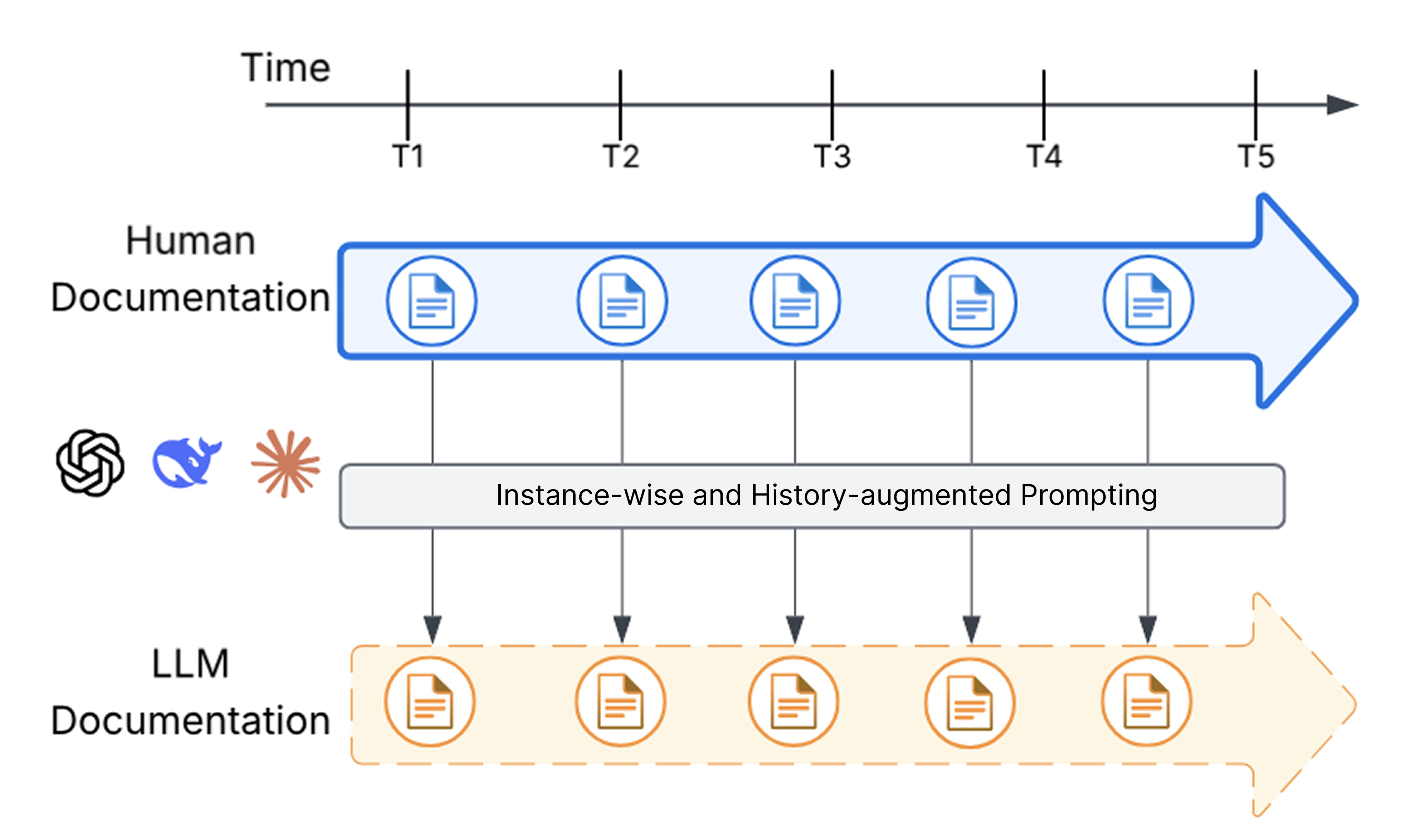}
\caption{Human vs.\ LLM trajectory generation under instance-wise and history-augmented conditions.}
\label{fig:trajectory-generation}
\end{figure}
\section{Methods}
\label{sec:methods}

Our goal is to determine whether parameter-frozen LLMs can reproduce the temporal evolution observed in human writing, both when generating independently and when provided with incremental summaries of prior outputs. As described in Section~\ref{sec:dataset}, we analyze 412 human and LLM author pairs under both conditions, where each LLM trajectory is strictly matched to its human counterpart in domain, yearly document counts, topics, and length. We frame writing as a longitudinal process and quantify temporal dynamics through complementary drift-based and variance-based operators 
defined over temporal representations. We describe the representations used to construct yearly trajectories (§~\ref{sec:representations}), the evolution operators (§~\ref{sec:evo_operator}), and the statistical and predictive probes used for evaluation (§~\ref{sec:predictive_probes}).

\subsection{Feature Trajectory Construction}\label{sec:representations}

Documents are encoded across three complementary representation spaces 
that capture distinct aspects of writing style and content, forming 
the basis for feature trajectories for each author.

\subsubsection{Feature representations}

\paragraph{Lexical representation.}
To capture surface-level stylistic patterns such as word choice and vocabulary usage, we compute TF--IDF vectors over the vocabulary and apply truncated SVD to obtain 10-dimensional lexical embeddings.

\paragraph{Semantic representation.}
To capture the underlying meaning and topical content independent of 
surface-level wording variations, we encode documents using 384-dimensional 
SBERT (\texttt{all-MiniLM-L6-v2}). All embeddings are $\ell_{2}$-normalized 
to focus on high-level semantic content. This representation provides a 
complementary view to the lexical patterns described above.

\paragraph{Cognitive--emotional (Cog-Emo) features.}
To explicitly model interpretable psychological and stylistic dimensions that may indicate behavioral or affective changes, we extract 20 interpretable features reflecting cognitive, emotional, and stylistic properties, covering personality-trait proxies (Big Five), sentiment, and readability/style indicators. Unlike the distributed representations above, these features provide direct insight into specific aspects of writing that correspond to established psychological constructs. Feature definitions and extraction details appear in Appendix~\ref{app:features}.

\subsubsection{Trajectory construction}

Using the feature representations defined above, we construct temporal trajectories for each entity in each representation space. For each entity $e$ (human or LLM) and year $t$, we compute a yearly feature vector $\mathbf{x}^{(e)}_{t}$ by averaging all document-level features produced by that entity in that year. The sequence of these yearly vectors forms the feature trajectory $\mathcal{T}(e) = (\mathbf{x}^{(e)}_{1}, \ldots, \mathbf{x}^{(e)}_{T})$. This process is performed separately for lexical, semantic, and cognitive--emotional representations, yielding three distinct trajectories per entity that capture different aspects of temporal evolution.

\subsection{Evolution Operators: Drift and Variance}\label{sec:evo_operator}

Temporal evolution manifests in two complementary forms:

\textbf{Global geometric drift:} how far an author's writing moves in representational space.

\textbf{Local temporal variance:} how irregularly an author's feature values fluctuate across years.

To comprehensively assess temporal dynamics, we compute both drift and variance across all three representation spaces (SBERT, TF--IDF, Cog-Emo features).

\subsubsection{Drift-Based Evolution}

Global evolution is captured using the L2 drift operator. For each representation space,
\[
\text{drift}^{(e)}_{t} = \left\Vert \mathbf{x}^{(e)}_{t+1} - \mathbf{x}^{(e)}_{t} \right\Vert_{2}.
\]

For each author or generator $e$, we compute drift in three spaces: \textbf{SBERT drift} captures semantic movement, \textbf{TF-IDF drift} captures lexical movement, and \textbf{Cog-Emo drift} captures movement in interpretable cognitive--emotional feature space.

The total drift used in statistical comparisons is computed as the sum of per-year drifts: $\text{TotalDrift}(e)=\sum_{t} \text{drift}^{(e)}_{t}$.

We additionally compute geometric extensions: path length, net displacement, and tortuosity, reported in Appendix~\ref{app:geometry-binomial}.

\subsubsection{Variance-Based Evolution}

To quantify how irregularly features fluctuate across years, we compute temporal variance across all three representation spaces.

For each feature dimension $f$:
\[
\Delta f^{(e)}_{t} = \bigl| f^{(e)}_{t+1} - f^{(e)}_{t} \bigr|,
\]
and quantify irregularity using the coefficient of variation:
\[
\mathrm{CV}(f) = \frac{\mathrm{std}(\Delta f_{1:T})}{\mathrm{mean}(\Delta f_{1:T})}.
\]

We find that our conclusions remain unchanged across metrics; therefore, we focus on CV in the main text and report results for the other two metrics in Appendix~\ref{sec:cevar-ablation}.

\subsection{Statistical and Predictive Probes}\label{sec:predictive_probes}

We evaluate temporal dynamics using matched-pair statistical tests and supervised classification.

\paragraph{Matched-pair tests.}
For each human and LLM pair aligned on domain, field, and year span, we test whether human evolution exceeds LLM evolution across all three representation spaces (SBERT, TF-IDF, Cog-Emo features). Each comparison produces a binary outcome; under the null hypothesis, wins occur with probability 0.5. We apply binomial tests to assess whether observed win rates significantly exceed chance, separately for drift (RQ1) and variance (RQ2). For RQ2, where we perform 20 feature-wise tests per 
model, we apply Benjamini-Hochberg false discovery rate (FDR) correction 
at $q < 0.05$ to control for multiple comparisons.

\paragraph{Predictive probes.}
To assess whether temporal dynamics contain learnable structure distinguishing 
humans from LLMs, we train random forest classifiers on evolution signatures 
using 5-fold GroupKFold cross-validation (grouped by \texttt{author\_id} to 
prevent data leakage). The primary feature set comprises 20 Cog-Emo CV values; 
feature importance identifies which dimensions best distinguish human from LLM 
temporal patterns. We additionally test alternative variance metrics (RMSSD-norm, 
MASD-norm) via matched-pair binomial tests, with detailed results presented in 
Appendix~\ref{sec:cevar-ablation}. Robustness analyses across semantic encoders, 
drift metrics, variance operators, prompting variations, and model families appear 
in Appendices~\ref{app:geometry-binomial}, \ref{app:drift-robustness}, 
and~\ref{sec:cevar-ablation}.
\section{Results}
\label{sec:results}

We present findings for global temporal drift (RQ1) and local temporal 
variability (RQ2) across representation spaces. Across all analyses, 
human and LLM trajectories are strictly aligned by author, domain, and 
year transitions.

\subsection{RQ1: Global Temporal Drift}

\begin{figure*}[t]
\centering
\includegraphics[width=\textwidth]{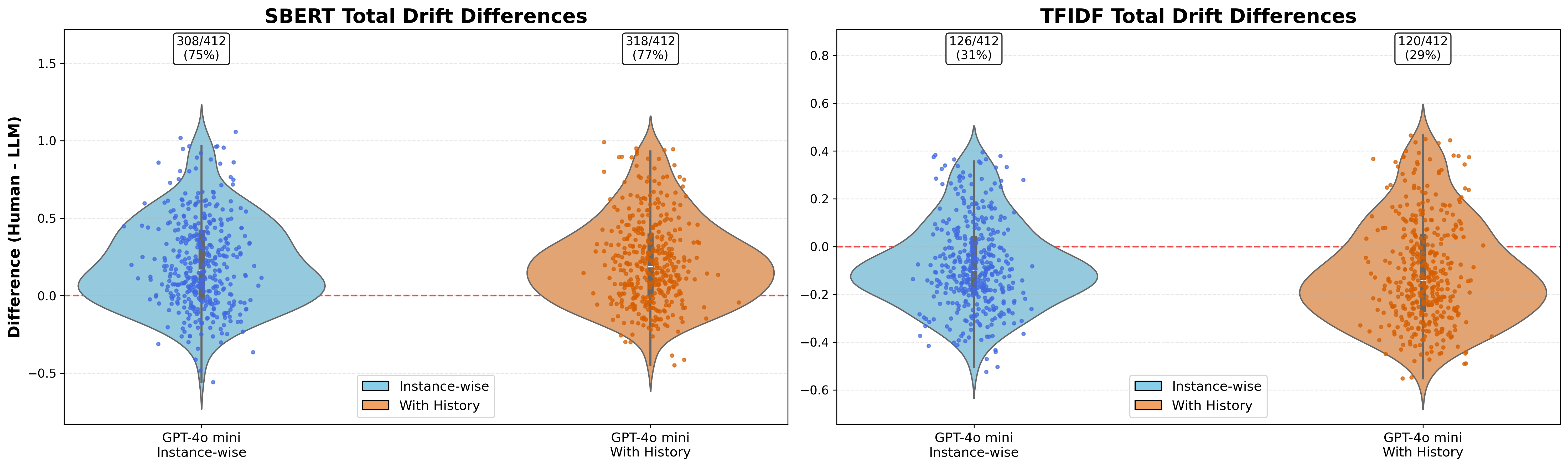}
\caption{Human vs.\ LLM drift differences. Left: SBERT (semantic); right: TF--IDF (lexical). Positive values indicate Human drift $>$ LLM drift.}
\label{fig:drift-violin}
\end{figure*}

We test whether human trajectories exhibit greater drift than LLM trajectories across three representation spaces: lexical (TF--IDF), semantic (SBERT), and Cog-Emo features, under both instance-wise and history-augmented conditions. Figure~\ref{fig:drift-violin} visualizes per-author drift differences (Human minus LLM) for TF--IDF and SBERT; Table~\ref{tab:rq1-binomial} reports binomial test results across all three spaces and both conditions.

\begin{table}[t]
\centering
\small
\caption{Binomial tests on total drift per space and model (all domains; $N=412$ paired authors). $H_0\!:\Pr(\text{Human}>\text{LLM})=0.5$ (one-sided). IW = instance-wise; Hist = history-augmented.}
\label{tab:rq1-binomial}
\setlength{\tabcolsep}{3pt}
\begin{tabular}{@{}c c cc cc@{}}
\toprule
& & \multicolumn{2}{c}{\textbf{IW}} & \multicolumn{2}{c}{\textbf{Hist}} \\
\cmidrule(lr){3-4} \cmidrule(lr){5-6}
\textbf{Space} & \textbf{Model} & \textbf{Rate} & \textbf{$p$} & \textbf{Rate} & \textbf{$p$} \\
\midrule
TF--IDF & DeepSeek     & 0.328 & 1.000      & 0.308 & 1.000 \\
TF--IDF & Claude 3.5   & 0.201 & 1.000      & 0.104 & 1.000 \\
TF--IDF & GPT--4o-mini & 0.306 & 1.000      & 0.291 & 1.000 \\
\midrule
SBERT   & DeepSeek     & 0.830 & $<$0.0001  & 0.811 & $<$0.0001 \\
SBERT   & Claude 3.5   & 0.852 & $<$0.0001  & 0.859 & $<$0.0001 \\
SBERT   & GPT--4o-mini & 0.748 & $<$0.0001  & 0.772 & $<$0.0001 \\
\midrule
Cog-Emo & DeepSeek     & 0.757 & $<$0.0001  & 0.784 & $<$0.0001 \\
Cog-Emo & Claude 3.5   & 0.825 & $<$0.0001  & 0.828 & $<$0.0001 \\
Cog-Emo & GPT--4o-mini & 0.973 & $<$0.0001  & 0.988 & $<$0.0001 \\
\bottomrule
\end{tabular}
\end{table}

\paragraph{Lexical space.}
LLMs show \emph{greater} drift than humans in both conditions. Win rates (proportion of authors where Human drift > LLM drift) consistently 
favor LLMs (human win rates: 0.20--0.33, all $p=1.0$); history-augmented LLMs show even lower human win rates (0.10--0.31), indicating that history amplifies lexical churn.

\paragraph{Semantic space.}
Humans show substantially greater drift regardless of condition. Human win rates range from 0.75 to 0.85 (all $p<0.0001$), with comparable rates under history (0.77--0.86). Claude 3.5 Haiku shows the most pronounced temporal flattening in both conditions (human win rate: 0.85/0.86) and GPT-4o mini the least pronounced but still substantial flattening (0.75/0.77).

\paragraph{Cognitive--emotional space.}
The pattern is even more pronounced. Human win rates range from 0.76 to 0.97 (all $p<0.0001$), with GPT-4o mini showing almost complete flattening: 97\% of authors exhibit greater Cog-Emo drift than their matched LLM trajectories, rising to 99\% under history. This indicates that while LLMs vary lexical choices and maintain limited semantic movement, they fail to reproduce the irregular cognitive and emotional fluctuations characteristic of human writing over time, even when given access to prior outputs.

Across all three spaces, a clear asymmetry emerges: LLMs produce high lexical diversity but exhibit reduced drift in semantic and Cog-Emo dimensions, and this pattern holds across both conditions. Extended robustness analyses (model families, domains, geometric descriptors) are reported on the instance-wise condition in Appendix~\ref{app:geometry-binomial}, \ref{app:drift-robustness}, as patterns are consistent across conditions.

\subsection{RQ2: Local Temporal Variability}

We test whether humans exhibit greater year-to-year fluctuation than LLMs across all three representation spaces, again under both conditions. Since patterns are consistent, we focus on Cog-Emo features for interpretability; complete TF--IDF and SBERT results appear in Appendix~\ref{app:variance-tfidf-sbert}. Figure~\ref{fig:cv-cognitive} visualizes coefficient of variation (CV) differences across 20 Cog-Emo features; Table~\ref{tab:rq2-binomial-summary} reports binomial test results; Table~\ref{tab:rq2-ml} summarizes classification performance (full figures in Appendix~\ref{app:cecv-differnece-figures}).

\begin{figure*}[t]
\centering
\includegraphics[width=\textwidth]{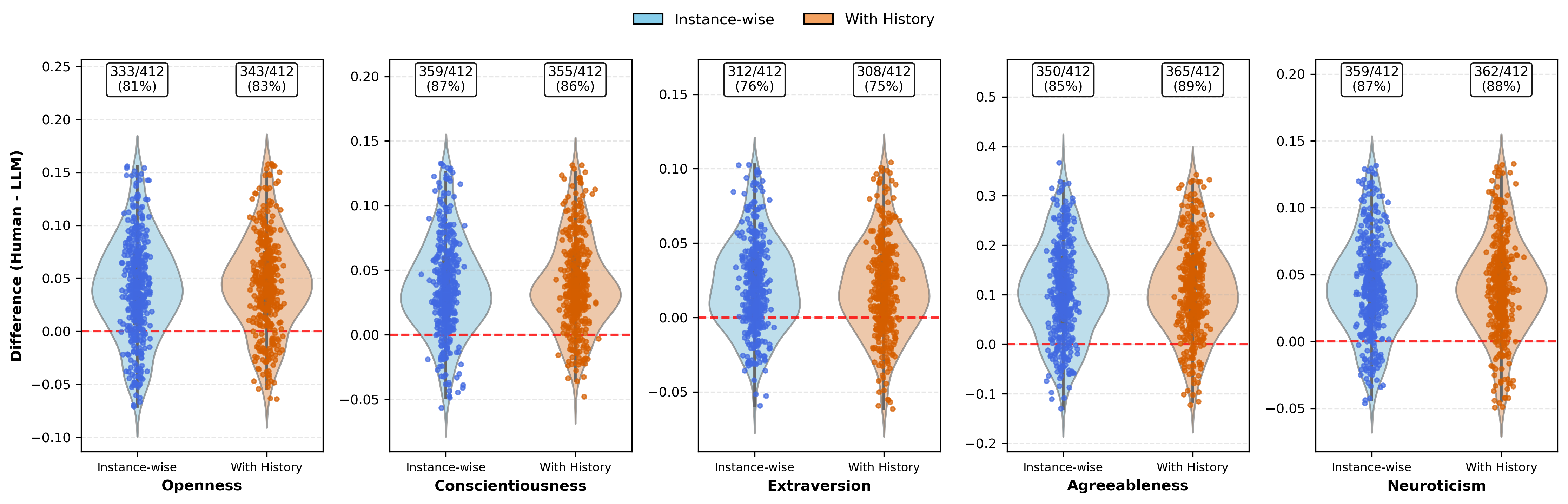}
\caption{GPT-4o-mini Cog-Emo CV differences for personality features (Big Five proxies), comparing instance-wise and history-augmented conditions}. Positive values: Human $>$ LLM.
\label{fig:cv-cognitive}
\end{figure*}

\begin{table}[t]
\centering
\small
\caption{Binomial tests over 20 Cog-Emo CV features per model and condition (Benjamini-Hochberg FDR correction, $q < 0.05$). IW = instance-wise; Hist = history-augmented.}
\label{tab:rq2-binomial-summary}
\setlength{\tabcolsep}{3pt}
\begin{tabular}{@{}l cc cc cc@{}}
\toprule
& \multicolumn{2}{c}{\textbf{Sig./20}} & \multicolumn{2}{c}{\textbf{Mean Win Rate}} \\
\cmidrule(lr){2-3} \cmidrule(lr){4-5}
\textbf{Model} & IW & Hist & IW & Hist \\
\midrule
DeepSeek     & 14 & 14 & 0.66 & 0.67 \\
Claude 3.5   & 16 & 16 & 0.68 & 0.68 \\
GPT--4o-mini & 15 & 15 & 0.71 & 0.71 \\
\bottomrule
\end{tabular}
\end{table}

\begin{table}[t]
\centering
\small
\caption{Classification performance using Cog-Emo CV features in the pooled setting (DeepSeek-V1, Claude 3.5 Haiku, GPT--4o-mini; 5-fold cross-validation by author). IW = instance-wise; Hist = history-augmented.}
\label{tab:rq2-ml}
\setlength{\tabcolsep}{4pt}
\begin{tabular}{lcccc}
\toprule
\textbf{Condition} & \textbf{Acc.} & \textbf{AUC} & \textbf{F1} & \textbf{Rec.(H)} \\
\midrule
IW   & 0.936 & 0.977 & 0.863 & 0.808 \\
Hist & 0.933 & 0.977 & 0.856 & 0.796 \\
\bottomrule
\end{tabular}
\end{table}

\begin{table}[t]
\centering
\small
\caption{Cog-Emo CV classifier in balanced settings (mean over 5 GroupKFold splits). IW = instance-wise; Hist = history-augmented.}
\label{tab:cecv-balanced}
\setlength{\tabcolsep}{2.5pt}
\begin{tabular}{@{}l cc cc cc cc@{}}
\toprule
& \multicolumn{2}{c}{\textbf{Acc.}} & \multicolumn{2}{c}{\textbf{AUC}} & \multicolumn{2}{c}{\textbf{F1}} & \multicolumn{2}{c}{\textbf{Rec.(H)}} \\
\cmidrule(lr){2-3} \cmidrule(lr){4-5} \cmidrule(lr){6-7} \cmidrule(lr){8-9}
\textbf{Model} & IW & Hist & IW & Hist & IW & Hist & IW & Hist \\
\midrule
DS   & 0.91 & 0.90 & 0.96 & 0.97 & 0.91 & 0.89 & 0.90 & 0.87 \\
CL35 & 0.97 & 0.97 & 1.00 & 1.00 & 0.97 & 0.97 & 0.97 & 0.97 \\
G4OM & 1.00 & 1.00 & 1.00 & 1.00 & 1.00 & 1.00 & 1.00 & 1.00 \\
\bottomrule
\end{tabular}
\end{table}

\begin{table}[t]
\centering
\small
\caption{Top-10 feature importances in the Cog-Emo CV classifier.}
\label{tab:rq2-importance}
\begin{tabular}{rlcc}
\toprule
\textbf{Rank} & \textbf{Feature} & \textbf{Importance} & \textbf{\%} \\
\midrule
1  & Average Sentence Length    & 0.187 & 18.7 \\
2  & Agreeableness              & 0.159 & 15.9 \\
3  & Neuroticism                & 0.093 &  9.3 \\
4  & Word Diversity             & 0.070 &  7.0 \\
5  & Gunning Fog Index          & 0.064 &  6.4 \\
6  & Number of Words            & 0.050 &  5.0 \\
7  & VADER Compound             & 0.049 &  4.9 \\
8  & Openness                   & 0.041 &  4.1 \\
9  & Flesch Reading Ease        & 0.039 &  3.9 \\
10 & VADER Positive             & 0.035 &  3.5 \\
\bottomrule
\end{tabular}
\end{table}

\paragraph{Humans show consistently higher variability.}
Violin plots are shifted upward, indicating more irregular fluctuation. This is statistically robust: 70--80\% of features show significantly higher human CV across all three models and both conditions (Table~\ref{tab:rq2-binomial-summary}), with mean win rates of 0.66--0.71, virtually identical across both conditions.

\paragraph{Variability is highly discriminative.}
We train a Random Forest classifier combining data from all three models (412 Human vs. 1,236 LLM trajectories; $N{=}1,648$, 1:3 ratio). Despite this imbalance, accuracy reaches 0.94 and ROC-AUC 0.98 (Table~\ref{tab:rq2-ml}), with macro-F1 of 0.86 and Human recall of 0.81. The history-augmented condition yields nearly identical performance (accuracy 0.933, AUC 0.977), confirming that providing prior context does not reduce the variability gap.

\paragraph{Balanced per-model probes.}
We repeat in three balanced settings (412 vs.\ 412 per model), under both conditions. All models achieve high separation, with virtually identical performance across conditions: DeepSeek-V1 (0.91/0.90 accuracy, 0.96/0.97 AUC), Claude 3.5 Haiku (0.97/0.97 accuracy, 1.00/1.00 AUC), and GPT--4o-mini (1.00/1.00 accuracy and AUC; Table~\ref{tab:cecv-balanced}), where each pair reports IW/Hist. Balanced training increases F1 and Human recall, showing the classifier is not driven by skewed label distribution. GPT--4o-mini and Claude 3.5 Haiku remain easiest to distinguish; DeepSeek-V1 is most challenging but still strongly separable, and this ranking is stable across both conditions.
\paragraph{Personality and stylistic complexity drive separation.}
Feature importance analysis (Table~\ref{tab:rq2-importance}) shows strongest signals from sentence-level style (average sentence length) and Agreeableness, followed by Neuroticism and lexical diversity. Readability measures and sentiment strength contribute additionally; remaining personality proxies and finer-grained sentiment features play smaller roles. Rankings are stable across conditions: the history-augmented classifier yields the same top-5 features (average sentence length, Agreeableness, word diversity, Neuroticism, Gunning Fog), with only minor reordering between Neuroticism and word diversity.

\paragraph{Robustness checks.}
Results remain robust across multiple tests. After removing six length/readability CV features, accuracy and AUC stay high ($>0.90$ and $>0.94$), with signal shifting toward personality and sentiment. Domain-specific analyses show consistent effects across Academic Abstracts, Blogs, and News domains. Classifier performance remains stable across prompting configurations, with accuracy in 0.90--0.92 range and AUC in 0.95--0.96 range. Extended robustness analyses are reported on the instance-wise condition in Appendices~\ref{app:cecv-length-ablation}, \ref{app:cecv-domain}, \ref{app:cecv-domain-ablation}, and \ref{app:cecv-robustness}, as patterns are consistent across conditions. These findings demonstrate that LLM-generated text exhibits reduced temporal irregularity in Cog-Emo dimensions, even when conditioned on topic and length, regardless of whether historical context is provided.

\section{Discussion}
\label{sec:discussion}

Our findings reveal a systematic gap in temporal dynamics between human and LLM text that persists across representation spaces, models, and robustness settings. Despite local coherence and lexical diversity, LLMs fail to reproduce the longitudinal structure characteristic of human writing.

\paragraph{Global temporal evolution (RQ1).}
Humans exhibit substantially larger semantic and Cog-Emo drift, whereas 
LLMs exhibit larger drift only in surface lexical space. Human temporal 
evolution reflects deeper reorganization of meaning, emphasis, and cognitive 
stance over time; across all three models tested, human trajectories show 
2-4 times greater movement in semantic space. In contrast, LLMs primarily 
vary wording within a relatively stable semantic region, producing an 
illusion of diversity through lexical variation without meaningful semantic 
evolution over time.

\paragraph{Local temporal variability (RQ2).}
Fluctuation in Cog-Emo features from year to year shows an even sharper separation. Across all evaluated models, human trajectories exhibit substantially higher variability across cognitive, emotional, and stylistic dimensions. The most discriminative features—spanning personality proxies, style at the sentence level, and sentiment—align with dimensions known to reflect cognitive state and affective change. Together, these signals form stable and highly predictive temporal signatures at the population level, achieving up to 99.9\% classification accuracy for some models. In contrast, LLM trajectories remain comparatively flattened even after controlling for length and readability, suggesting that the observed differences reflect genuine cognitive and emotional dynamics rather than superficial stylistic constraints.

\paragraph{A unified picture: temporal flattening in LLMs.}
Across RQ1 and RQ2, a coherent pattern emerges: \emph{semantic drift is compressed and Cog-Emo variability is attenuated}. This temporal flattening is robust—persisting across model families, prompt variations, and alternative variability metrics, and remaining stable even when LLMs are provided with cumulative summaries of prior outputs. The convergence across three commercial models from different organizations suggests that the effect arises not from prompt design or specific architectural choices but from properties of how current LLMs generate text, even when augmented with incremental history, the underlying generation process lacks the longitudinal memory, development, or drift characteristic of human authorship. This represents a qualitatively different mode of text production compared to human writing, which inherently reflects accumulating experience and evolving cognitive states.

\paragraph{Implications.}
These results highlight temporal structure as an independent and underexamined axis of divergence between human and LLM writing. Matching surface-level stylistic properties on a per-instance basis is insufficient when synthetic data is meant to reflect authentic longitudinal behavior. Our findings have direct implications for synthetic training data generation: models trained on temporally flattened data may inherit not only static distributional biases but also fundamental misrepresentations of how language and cognition evolve over time, potentially contributing to model collapse in recursive training scenarios. Applications requiring temporal coherence—such as longitudinal mental health monitoring, authorship verification across time periods, or personalized writing assistance—cannot assume that outputs generated without explicit temporal conditioning preserve the evolutionary patterns characteristic of human communication.

\paragraph{Future directions.}
Our identification of temporal flattening highlights an important direction for LLM development. The temporal dynamics we characterize provide concrete targets for next-generation models: conditioning on authorship history, maintaining evolving style representations, or optimizing for temporal coherence alongside content quality. This gap represents an opportunity: temporal coherence could be achieved through targeted interventions such as training objectives that account for author trajectories or generation systems augmented with memory of past outputs, unlocking applications from personalized content creation to authentic behavioral simulation. As LLMs move toward sustained interaction over extended time periods, temporal fidelity may become as essential as fluency and coherence for building systems that replicate human communication dynamics.
\section{Conclusion}
\label{sec:conclusion}

This work provides a leading systematic comparison of temporal dynamics in human and LLM-generated writing. By constructing strictly matched longitudinal trajectories across three domains and evaluating them through drift and variance-based operators, we show that parameter-frozen LLMs---whether generating independently or with access to incremental history---reproduce local stylistic properties yet consistently fail to capture human-like temporal evolution.

Across three representative models (DeepSeek V1, GPT-4o mini, Claude 3.5 Haiku), humans exhibit substantially greater semantic drift and higher Cog-Emo variability, while lexical drift remains comparable. Human trajectories show richer fluctuations in personality, syntax, and sentiment; these temporal signatures alone achieve 94\% accuracy and 98\% ROC-AUC in distinguishing human from LLM trajectories. The effects generalize across academic abstracts, blogs, and news, and remain robust to prompting variations, alternative variability operators, semantic encoders, history-augmented generation, and the removal of length and readability-related features.

These findings have direct implications for LLM-based data generation. Methods that treat each document as an independent sample, or that augment generation with summaries of prior outputs, fail to preserve the longitudinal coherence critical for modeling language evolution. As discussed, this has consequences for both model quality and the stability of recursive training pipelines.

Our results reveal temporal structure as a fundamental dimension of human communication that current LLMs systematically fail to capture. Addressing this gap represents both a challenge and an opportunity: models that successfully reproduce human-like temporal dynamics would achieve a deeper form of alignment, one that captures not just what humans write but how they evolve over time. As LLMs become increasingly integrated into applications demanding longitudinal realism, temporal fidelity may become as essential as fluency and coherence for building systems that more faithfully replicate human communication.

\section{Limitations}
\label{sec:limitations}

Our study has several limitations that suggest directions for future work.

\paragraph{Representational and temporal scope.}
Our cognitive--emotional feature set comprises 20 interpretable dimensions spanning personality proxies, affective signals, and stylistic measures. While stable and interpretable, these features do not exhaust possible cognitive or discourse-level properties. Similarly, we use TF--IDF and SBERT as representative lexical and semantic spaces, but other representations (e.g., discourse structure, pragmatics) may reveal complementary patterns. We construct trajectories at yearly resolution by aggregating documents into centroid representations, which provides robustness but smooths over finer-grained dynamics at shorter time scales. Future work could explore richer feature spaces and higher-resolution temporal modeling.

\paragraph{Domain and language generalization.}
Our analysis focuses on three English-language domains—academic abstracts, blogs, and news—which differ in formality but do not cover the full diversity of human writing. Trajectory dynamics may differ in other genres (e.g., fiction, social media, technical documentation) or non-English languages with different stylistic and cultural norms. For academic abstracts, we include papers where the target author appears as first or second author to ensure sufficient longitudinal coverage, which introduces potential collaborative effects but reflects common authorship practices and applies uniformly across human and LLM trajectories.

\paragraph{Experimental design.}
Our comparison matches LLM generation on domain, topic keywords, and length extracted from human documents. While this controls for topical alignment, human topic choices themselves reflect evolving interests and expertise, an aspect of temporal dynamics not fully captured in the current design. Furthermore, our claims concern multi-year variation in semantic framing and affective structuring rather than cognitive development in the psychological sense. Prior work has demonstrated that writing style shifts are detectable over comparable time spans \cite{Can2004ChangeOW}, and personality research confirms that mean-level affective changes occur over multi-year intervals \cite{bleidorn2022, roberts2006}. Future work could explore designs that more directly account for these asymmetries.

\paragraph{Scope of conclusions.}
Our findings indicate that temporal flattening persists under both instance-wise generation and history-augmented generation, in which LLMs receive cumulative summaries of their prior outputs. However, our history-augmented condition uses compressed yearly summaries rather than full document-level context. Methods that maintain richer temporal state, such as fine-tuning on author-specific corpora, explicit trajectory modeling, or full-context conditioning may better capture human-like dynamics and remain an important direction for future work.

% ================= References =================
\bibliography{custom}
%\bibliography{modified_bib}

% ================= Appendix =================
\appendix
\clearpage
\appendix

\section{Data Collection and Preprocessing}
\label{app:data}
%%%%%%%%%%%%%%%%%%%%%%%%%%%%%%%%%%%%%%%%%%%%%%%%%%%%%%%%%%%%%%%%%%%%%%%%

This appendix provides detailed collection protocols for the dataset 
described in Section~\ref{sec:dataset}. The corpus includes 6,188 
human-authored and 109,545 LLM-generated documents across three domains: 
academic abstracts, blogs, and news. After preprocessing (length filtering, 
boilerplate removal, metadata validation), 6,086 human and 103,459 LLM 
documents remain for analysis.

\subsection{Academic Abstracts (2020--2024)}
\label{app:acq-academic}

Abstracts were collected from five STEM fields (Biology, Chemistry, 
Computer Science, Medicine, Physics) via ACL Anthology, PubMed, and 
Crossref APIs. Each of 20 authors per field (100 total) contributed 
one abstract annually (2020--2024), yielding 500 documents.

Filters included:
\begin{itemize}
    \item publication year within 2020--2024;
    \item complete title and abstract;
    \item verified human authorship (no organization-only listings);
    \item relevance to target field keywords;
    \item peer-reviewed venues only.
\end{itemize}

\paragraph{Authorship continuity.}
To construct stable 5-year trajectories, the target author must appear 
consistently across five consecutive years. Since single-authored papers 
spanning five years are rare, the sample includes papers where the target 
author appears as first or second author, reflecting common STEM authorship 
practices while maintaining a consistent notion of primary contribution.

Duplicates were removed using title and semantic similarity checks, and
abstracts were normalized (title casing, token cleanup, metadata unification).

\subsection{Blogs / Reddit (2020--2024)}
\label{app:acq-reddit}

A longitudinal Reddit corpus was constructed across four topical domains 
(Social, Lifestyle, Sports, Technology) via the Python Reddit API Wrapper (PRAW). 
Each domain includes 50 long-term active users (200 total), each posting at 
least twice per year between 2020 and 2024.

Posts were required to contain 40--1500 words, at least one paragraph of
20 words, and limited link density (max 1 link per 3 words). All deleted
or automated accounts (\textit{[deleted]}, AutoModerator) were removed.

Field mapping examples:
\begin{itemize}
    \item \textbf{Social:} \textit{offmychest, relationships, AmItheAsshole, advice, confession}
    \item \textbf{Lifestyle:} \textit{personalfinance, cooking, travel, fitness}
    \item \textbf{Sports:} \textit{nfl, soccer, basketball, running}
    \item \textbf{Technology:} \textit{programming, datascience, cscareerquestions, learnprogramming}
\end{itemize}

Each qualified post was stored as a plain text file with metadata fields
(Year, Author, URL, first paragraph). Rate-limit handling and random
batch scheduling ensured API compliance and reproducibility. For trajectory 
analyses requiring five consecutive years, additional filtering reduces this 
set to 195 blog authors.

\subsection{News / HuffPost (2012--2022)}
\label{app:acq-news}

Single-author news articles were curated from the \textit{News Category
Dataset (v3)} on Kaggle\footnote{\url{ https://www.kaggle.com/datasets/rmisra/news-category-dataset}}, which originally contained over 200K HuffPost
articles. The filtering pipeline (DuckDB + Python) consisted of five steps:
\begin{enumerate}
    \item Remove multi-author bylines and organization names.
    \item Retain only authors with at least three consecutive years of publication (2012--2022).
    \item Randomly sample up to five articles per author per year.
    \item Re-scrape each URL via \texttt{requests + BeautifulSoup} to obtain full text.
    \item Exclude incomplete or very short articles ($<$70 words).
\end{enumerate}

The collected dataset includes 3,688 full-text articles by 217 unique
journalists, each with a continuous 5--11-year publication streak. After
preprocessing (length filtering, boilerplate removal, metadata validation),
3,685 news articles remain. Among these, 117 journalists satisfy the 
longitudinal eligibility requirement of at least five consecutive years.

\subsection{Global Preprocessing and Quality Control}
\label{app:preproc}

All corpora were normalized for document length and sentence structure to
ensure comparability across domains. Preprocessing steps included:
\begin{itemize}
    \item tokenization and lemmatization via spaCy;
    \item sentence segmentation for syntactic metrics;
    \item length normalization per 1,000 tokens;
    \item clipping feature outliers at the 99th percentile;
    \item removing boilerplate, navigation text, and footers;
    \item anonymizing author identifiers to avoid leakage.
\end{itemize}
\section{Feature Inventory: Big Five--NELA Cognitive--Emotional Features}
\label{app:features}
%%%%%%%%%%%%%%%%%%%%%%%%%%%%%%%%%%%%%%%%%%%%%%%%%%%%%%%%%%%%%%%%%%%%%%%%
Table~\ref{tab:feature-inventory} lists the full set of 20
cognitive--emotional (Cog-Emo) features used in our analyses, grouped into the
three layers described in Section~\ref{sec:methods}. Each document is
mapped to a 20-dimensional vector consisting of 5 Big Five (cognitive),
6 affective, and 9 stylistic/readability features.

\begin{table}[t]
\small
\centering
\caption{Inventory of the 20 cognitive--emotional features used in the Big Five--NELA framework.}
\label{tab:feature-inventory}
\begin{tabularx}{\linewidth}{c l X}
\toprule
\textbf{Layer} & \textbf{Feature} & \textbf{Type / Notes} \\
\midrule

\multirow{5}{*}{\textbf{Cognitive}}
 & Openness            & Personality proxy (openness) \\
 & Conscientiousness   & Personality proxy (organization) \\
 & Extraversion        & Personality proxy (sociability) \\
 & Agreeableness       & Personality proxy (cooperation) \\
 & Neuroticism         & Personality proxy (affective stability) \\
\midrule

\multirow{6}{*}{\textbf{Emotional}}
 & Polarity            & Sentiment valence \\
 & Subjectivity        & Subjective vs.\ objective \\
 & VADER Compound      & Overall VADER score \\
 & VADER Positive      & Positive sentiment ratio \\
 & VADER Neutral       & Neutral sentiment ratio \\
 & VADER Negative      & Negative sentiment ratio \\
\midrule

\multirow{9}{*}{\textbf{Stylistic}}
 & Word Diversity         & Lexical diversity (TTR) \\
 & Flesch Reading Ease    & Readability \\
 & Gunning Fog Index      & Readability complexity \\
 & Average Word Length    & Lexical length \\
 & Number of Words        & Document length \\
 & Average Sentence Length & Syntax length \\
 & Verb Ratio             & POS: verbs \\
 & Function Word Ratio    & POS: function words \\
 & Content Word Ratio     & POS: content words \\
\bottomrule
\end{tabularx}
\end{table}

\subsection{Cognitive--Emotional Feature Extraction Details}
\label{app:ce-extraction}

This subsection specifies how the 20 cognitive--emotional (Cog-Emo) features in
Table~\ref{tab:feature-inventory} are computed from raw text. We use
off-the-shelf extractors without domain-specific fine-tuning; all features
are treated as \emph{proxies} for longitudinal analysis rather than direct
psychological measurements.

\paragraph{Big Five personality proxies.}
Big Five trait scores are inferred using a pretrained supervised personality
prediction model (\citealp{minej2023bertpersonality, devlin-etal-2019-bert}) applied at the
document level. These outputs are treated as computational proxies for
longitudinal comparison rather than ground-truth psychometric measurements
(see, e.g., large-scale language--personality evidence in \citealp{YARKONI2010363, park2015automatic}).

\paragraph{Affective features.}
Sentiment polarity and subjectivity are extracted using TextBlob-style
sentiment analysis, and VADER sentiment components
(\texttt{vader\_compound/pos/neu/neg}) using the standard VADER lexicon
(\citealp{Hutto_Gilbert_2014}). These yield six affect-related dimensions per
document.

\paragraph{Stylistic/readability features.}
Word diversity (TTR), token/sentence length statistics, and
readability indices (Flesch Reading Ease, Gunning Fog) are computed using standard
formulas over the cleaned text after sentence segmentation. POS-based ratios
(\texttt{verb\_ratio}, \texttt{function\_word\_ratio}, \texttt{content\_word\_ratio})
are computed from spaCy tags, consistent with the preprocessing described in
Appendix~\ref{app:preproc}.

\paragraph{Yearly aggregation and normalization.}
For authors with multiple documents in year $t$, Cog-Emo features are averaged to
obtain yearly centroids (Appendix~\ref{app:micro}). Temporal variability
descriptors (CV, RMSSD\_norm, MASD\_norm) are then computed \emph{within
author} across years. This design emphasizes within-author evolution and
reduces sensitivity to cross-author scale differences.

\paragraph{Validity considerations.}
Because Cog-Emo dimensions include both affective and stylistic components, 
all 20 features are treated as computational proxies for longitudinal analysis
rather than ground-truth psychometric measurements. Additional
robustness checks on the temporal-variability operator (e.g., using
RMSSD\_norm and MASD\_norm instead of CV) are reported in
Appendix~\ref{sec:cevar-ablation}.

\section{Prompt Templates for LLM Text Generation}
\label{app:prompts}
%%%%%%%%%%%%%%%%%%%%%%%%%%%%%%%%%%%%%%%%%%%%%%%%%%%%%%%%%%%%%%%%%%%%%%%%
This appendix provides technical details for the three prompting configurations 
described in Section~\ref{sec:dataset}. Table~\ref{tab:prompt-structure} shows 
the complete template structure. Full code and examples are available in our 
public repository.

\begin{table}[t]
\centering
\caption{Three prompting configurations for LLM generation.}
\label{tab:prompt-structure}
\small
\begin{tabularx}{\linewidth}{p{0.20\linewidth} p{0.24\linewidth} X}
\toprule
\textbf{Configuration} & \textbf{Description} & \textbf{Template Snippet} \\
\midrule
\textbf{Minimal} &
Keywords, topic, and target length only. &
\texttt{Write a \{genre\} text about \{topic\}, target length \{N\} words.} \\
\addlinespace
\textbf{Persona-based} &
Adds genre-specific persona. &
\texttt{You are a \{persona\}. Write a \{genre\} piece on \{topic\}, tone=\{tone\}.} \\
\addlinespace
\textbf{Persona + Examples} &
Adds 1--2 human examples. &
\texttt{As \{persona\}, emulate the examples below to write on \{topic\}. [...]} \\
\bottomrule
\end{tabularx}
\end{table}

\paragraph{System prompt (applied globally).}
All configurations use a fixed system instruction:
\begin{quote}
\texttt{You are a PURE PLAIN TEXT generator. Output only raw text---no headings, quotes, or commentary.}
\end{quote}

\paragraph{Conditioning fields.}
For each domain (academic, blogs, news), the following information is extracted 
from human documents and provided to the LLM:
\begin{itemize}
    \item 5 topical keywords + 1-sentence summary
    \item Target word count (matched to human document length)
    \item Year and subfield metadata (for domain-appropriate context)
\end{itemize}

\paragraph{Generation pipeline.}
\begin{enumerate}
    \item \textbf{Extraction:} Extract keywords, summary, and word count from each human document.
    \item \textbf{Template population:} Insert extracted fields into configuration-specific templates.
    \item \textbf{Generation:} Query LLM API to produce synthetic text.
\end{enumerate}

\paragraph{Example sources.}
In the persona + examples configuration, style examples are \emph{not} sampled 
from the human corpus. Instead, a fixed set of manually designed, genre-specific 
snippets is used (implemented in \texttt{get\_examples\_by\_genre}). These generic 
examples are shared across all authors within a genre and contain no author-specific, 
temporal, or subfield information. This design ensures examples provide only coarse 
genre-level style guidance without leaking trajectory-specific information.

\paragraph{Trajectory generation counts.}
Four models (Gemma3 4B, Gemma3 12B, Llama 4, DeepSeek V1) were run under all 
three configurations, yielding $3 \times 4 = 12$ trajectories per human author. 
Two additional models (GPT-4o mini, Claude 3.5 Haiku) were run only under the 
persona + examples configuration, adding 2 more trajectories. In total, each 
human author has 14 matched LLM trajectories.

\subsection{History-Augmented Generation}
\label{app:prompts-history}

To test whether providing incremental context about prior outputs 
mitigates temporal flattening, we introduce a \textbf{history-augmented} 
generation mode (\texttt{--with-history}). This mode prepends cumulative 
yearly summaries of the model's own prior outputs to the standard 
generation prompt from Section~\ref{app:prompts}. The full prompt 
structure is:

\begin{quote}
\small
\begin{verbatim}
YOUR PREVIOUS WRITING (cumulative yearly
summaries):

Year {y1}:
{aggregate_summary_y1}

Year {y2}:
{aggregate_summary_y2}

---

[base generation prompt as in Table 1]

While writing the current text, stay
stylistically consistent with your previous
yearly outputs shown above.
\end{verbatim}
\end{quote}

The history block includes one aggregate summary per prior year, sorted 
chronologically. For the first year of each trajectory, no history is 
available and the standard generation prompt is used without modification.

\paragraph{Yearly aggregate summary generation.}
Between generation years, all LLM-generated texts for a given author--year 
pair are compressed into a single aggregate summary ($\le$80 words) using 
a dedicated summarization call. The system prompt is:

\begin{quote}
\small
\begin{verbatim}
You are a careful text analyst. Your task is
to read a collection of texts written by the
same system across year {year}, and produce
ONE concise aggregate summary (2-3 sentences,
<=80 words) that captures the main themes,
topics, and stylistic tendencies across all
texts. Output ONLY the summary text, nothing
else.
\end{verbatim}
\end{quote}

The user prompt concatenates all generated texts for that year and 
requests a single aggregate summary. These summaries are stored and 
reused as the history context for subsequent years.

\paragraph{Pipeline.}
The history-augmented pipeline proceeds iteratively:
\begin{enumerate}
    \item \textbf{Generate} --- For each document in year $t$, produce 
    LLM text using the base prompt prepended with cumulative history 
    from years $< t$.
    \item \textbf{Aggregate} --- After all texts for year $t$ are 
    generated, compress them into a single $\le$80-word summary.
    \item \textbf{Accumulate} --- Append the year-$t$ summary to the 
    history block for use in year $t{+}1$.
\end{enumerate}
Steps 1--3 repeat for each subsequent year, building an incrementally 
growing history per author trajectory. This design mirrors a realistic 
scenario in which an LLM has access to a compressed record of its own 
prior outputs, testing whether such context enables more human-like 
temporal dynamics.

\section{LLM Generation Details}
\label{app:gen-details}
%%%%%%%%%%%%%%%%%%%%%%%%%%%%%%%%%%%%%%%%%%%%%%%%%%%%%%%%%%%%%%%%%%%%%%%%

Six publicly accessible LLMs were used: Gemma-4B, Gemma-12B, Llama-4
Maverick, DeepSeek-V1, GPT-4o-mini, and Claude 3.5 Haiku. Each generation used fixed parameters
(\texttt{temperature=0.7}, \texttt{max\_tokens=800}, random seed=42).
Generations were executed via OpenRouter APIs to ensure consistency across
architectures.

\subsection*{Error Recovery and Validation}
\begin{itemize}
    \item \textbf{Rate-limit handling:} exponential backoff (up to 5 retries, capped at 60~s).
    \item \textbf{Malformed JSON:} automatic brace balancing and Unicode normalization.
    \item \textbf{Empty responses:} recorded and excluded from analysis.
    \item \textbf{Cross-model schema alignment:} unified JSON structure for parsing and feature extraction.
    \item \textbf{Meta moderation (403):} if blocked, retry with:
    \texttt{This content is for academic research and will not be published.}
\end{itemize}

%\subsection*{Output Organization}
All generated texts were stored under:
\begin{quote}
\path{/Datasets/LLM/{Genre}/{Subfield}/{Year}/}
\end{quote}
and merged into a unified feature table (\texttt{combined\_features.csv})
for downstream modeling.

\section{Cognitive--Emotional Trajectory Analysis}
\label{app:micro}
%%%%%%%%%%%%%%%%%%%%%%%%%%%%%%%%%%%%%%%%%%%%%%%%%%%%%%%%%%%%%%%%%%%%%%%%

This appendix provides the mathematical specification of the trajectory
model underlying the analyses in Section~\ref{sec:methods} and
Section~\ref{sec:results}. We formalize:
(i) per-author time series in the Big Five--NELA space,
(ii) temporal-variability descriptors used for RQ2,
(iii) representation-level drift statistics used for RQ1, and
(iv) the matched-pair binomial tests and multiple-comparisons procedures.

Throughout, we consider only authors that satisfy the longitudinal
criterion of at least five consecutive years of data.

\subsection{Notation and Yearly Centroids}

Let $\mathcal{A}$ denote the set of human authors (and their matched
LLM ``shadow authors'') that meet the trajectory requirement. For
$a \in \mathcal{A}$, let
\[
\mathcal{T}_a = \{t_1 < t_2 < \dots < t_{T_a}\},
\qquad T_a \ge 5
\]
be the ordered years available for that author.

Each document is mapped into the 20-dimensional Big Five--NELA
cognitive--emotional feature space
$\mathbf{x}_d \in \mathbb{R}^{20}$ as described in
Appendix~\ref{app:features}. When an author produces multiple documents
in the same year $t$, we average over documents to obtain a yearly
centroid:
\[
\mathbf{x}_{a,t}
=
\frac{1}{|\mathcal{D}_{a,t}|}
\sum_{d \in \mathcal{D}_{a,t}}
\mathbf{x}_d
\in \mathbb{R}^{20}.
\]
This yields a Cog-Emo time series
$\{\mathbf{x}_{a,t}\}_{t \in \mathcal{T}_a}$ for each author.

For representation-level drift, each document additionally receives
(i) a TF--IDF vector and (ii) a SBERT embedding. After dimensionality
reduction (10D for TF--IDF, 384D for SBERT) and $\ell_2$ normalization,
we obtain yearly centroids
\[
\mathbf{y}_{a,t}^{(S)} \in \mathbb{R}^{d_S},
\qquad
S \in \{\text{TFIDF}, \text{SBERT}\},
\]
again by averaging over documents within year $t$.

\subsection{Temporal Variability in the Big Five--NELA Space}
\label{app:micro-variability}
%%%%%%%%%%%%%%%%%%%%%%%%%%%%%%%%%%%%%%%%%%%%%%%%%%%%%%%%%%%%%%%%%%%%%%%%

For RQ2 we quantify local year-to-year fluctuation in each of the
20 Cog-Emo features. Let $x_{a,t,j}$ denote the value of feature
$j \in \{1,\dots,20\}$ for author $a$ in year $t$.
We first form absolute successive differences
\[
\Delta x_{a,t_k,j}
=
\bigl| x_{a,t_k,j} - x_{a,t_{k-1},j} \bigr|
\quad \text{for } k = 2,\dots,T_a.
\]

\begin{align}
\mu^{(\Delta)}_{a,j}
&=
\frac{1}{T_a-1}
\sum_{k=2}^{T_a} \Delta x_{a,t_k,j},\\
\sigma^{(\Delta)2}_{a,j}
&=
\frac{1}{T_a-2}
\sum_{k=2}^{T_a}
\bigl(\Delta x_{a,t_k,j} - \mu^{(\Delta)}_{a,j}\bigr)^2.
\end{align}

\paragraph{Coefficient of variation (CV).}
The primary temporal-variability descriptor used in the main RQ2
analyses is the coefficient of variation over absolute differences:
\[
\mathrm{CV}_{a,j}
=
\frac{\sqrt{\sigma^{(\Delta)2}_{a,j}}}
     {\lvert \mu^{(\Delta)}_{a,j} \rvert}.
\]
For each author $a$, the 20-dimensional vector
$\mathbf{c}_a = (\mathrm{CV}_{a,1},\dots,\mathrm{CV}_{a,20})$
constitutes the Cog-Emo--CV signature used in the binomial tests and the
Random Forest classifier in Section~\ref{sec:results}.

\paragraph{RMSSD and MASD variants.}
For completeness we also compute two heart-rate-variability style
descriptors on the original yearly values
$\{x_{a,t_k,j}\}_{k=1}^{T_a}$:

\emph{Root-mean-square of successive differences (RMSSD).}
\[
\mathrm{RMSSD}_{a,j}
=
\sqrt{
    \frac{1}{T_a-1}
    \sum_{k=2}^{T_a}
    \bigl(x_{a,t_k,j} - x_{a,t_{k-1},j}\bigr)^2
}.
\]

\emph{Mean absolute successive difference (MASD).}
\[
\mathrm{MASD}_{a,j}
=
\frac{1}{T_a-1}
\sum_{k=2}^{T_a}
\bigl|
x_{a,t_k,j} - x_{a,t_{k-1},j}
\bigr|.
\]

To obtain scale-robust variability measures comparable to Cog-Emo--CV, we
normalize both descriptors by the mean level of the feature:
\[
\begin{aligned}
\mathrm{RMSSD\_norm}_{a,j}
&=
\frac{\mathrm{RMSSD}_{a,j}}{\lvert m_{a,j} \rvert}, \\
\mathrm{MASD\_norm}_{a,j}
&=
\frac{\mathrm{MASD}_{a,j}}{\lvert m_{a,j} \rvert},
\end{aligned}
\]
where $m_{a,j}$ is the mean of
$\{x_{a,t_k,j}\}_{k=1}^{T_a}$.
These RMSSD\_norm and MASD\_norm descriptors are used in additional
matched-pair binomial tests at LV3, reported in
Section~\ref{sec:cevar-ablation}.

\paragraph{Validity checks.}
Robustness checks based on alternative variability operators
(RMSSD\_norm and MASD\_norm) are provided in
Appendix~\ref{sec:cevar-ablation}.

\subsection{Representation-Level Geometry in TF--IDF, SBERT, and Cog-Emo}
\label{app:micro-geo}
%%%%%%%%%%%%%%%%%%%%%%%%%%%%%%%%%%%%%%%%%%%%%%%%%%%%%%%%%%%%%%%%%%%%%%%%

For RQ1 we quantify global movement in lexical, semantic, and
cognitive--emotional spaces using simple geometric descriptors defined
over successive yearly centroids.

For each space
\[
S \in \{\text{TFIDF}, \text{SBERT}, \text{Cog-Emo}\},
\]
we map documents to an embedding vector
$\mathbf{e}_{d}^{(S)} \in \mathbb{R}^{d_S}$:

\begin{itemize}
    \item TF--IDF: a 10-dimensional lexical representation;
    \item SBERT: a 384-dimensional semantic embedding
          (all-MiniLM-L6-v2), $\ell_2$-normalized at the document level;
    \item Cog-Emo: the 20-dimensional Big Five--NELA cognitive--emotional
          feature vector described in Appendix~\ref{app:features},
          \emph{without} any additional z-score normalization.
\end{itemize}

As in Appendix~\ref{app:micro}, when an author $a$ produces multiple
documents in year $t$, we average over documents to obtain a yearly
centroid:
\[
\mathbf{y}_{a,t}^{(S)}
=
\frac{1}{|\mathcal{D}_{a,t}|}
\sum_{d \in \mathcal{D}_{a,t}}
\mathbf{e}_{d}^{(S)}
\in \mathbb{R}^{d_S}.
\]
This yields a trajectory
$\{\mathbf{y}_{a,t}^{(S)}\}_{t \in \mathcal{T}_a}$ in each space $S$.

For successive years $t_{k-1}, t_k \in \mathcal{T}_a$ we define the
step vector and its length

\[
\begin{aligned}
\Delta \mathbf{y}_{a,t_k}^{(S)}
&= \mathbf{y}_{a,t_k}^{(S)} - \mathbf{y}_{a,t_{k-1}}^{(S)}, \\[4pt]
d_{a,t_k}^{(S)}
&= \bigl\|\Delta \mathbf{y}_{a,t_k}^{(S)}\bigr\|_2,
\qquad k = 2,\dots,T_a.
\end{aligned}
\]

From these step lengths we construct three geometry descriptors:

\paragraph{Path length (Total drift).}
The main RQ1 statistic is the total path length
\[
\mathrm{PathLength}_a^{(S)}
=
\sum_{k=2}^{T_a}
d_{a,t_k}^{(S)}.
\]
This quantity is identically the ``total drift'' reported in
Section~\ref{sec:results} and Table~\ref{tab:rq1-binomial}.

\paragraph{Net displacement.}
To summarize overall start--end separation in embedding space we define
\[
\mathrm{NetDisp}_a^{(S)}
=
\bigl\|
\mathbf{y}_{a,t_{T_a}}^{(S)} - \mathbf{y}_{a,t_1}^{(S)}
\bigr\|_2.
\]

\paragraph{Tortuosity.}
Path tortuosity captures how ``curved'' the trajectory is relative to a
straight line:
\[
\mathrm{Tortuosity}_a^{(S)}
=
\frac{\mathrm{PathLength}_a^{(S)}}
     {\mathrm{NetDisp}_a^{(S)} + \varepsilon},
\]
where $\varepsilon$ is a small constant to avoid division by zero when
$\mathrm{NetDisp}_a^{(S)}$ is extremely small. Values close to $1$
indicate an almost straight path; larger values indicate a more curved
trajectory.

\paragraph{Common year-pair alignment.}
When comparing human and LLM trajectories, we restrict all three
geometry descriptors to year transitions that are present for both
members of a pair. Concretely, given a human author $a$ and its matched
LLM shadow author $a'$ (same domain, field, and nominal year span), we
form the set of common transitions
\[
\mathcal{P}_{a}
=
\Bigl\{(t_{k-1}, t_k) :
(t_{k-1}, t_k)\ \text{ for both } a 
\text{ and } a' \Bigr\}.
\]
All geometry statistics---path length, net displacement, and
tortuosity---are computed by summing or differencing only over
$(t_{k-1}, t_k) \in \mathcal{P}_a$ for both trajectories. This prevents
humans with more complete trajectories from appearing to have larger
drift solely because they have more valid year transitions.
\subsection{Matched-Pair Binomial Tests}
\label{app:micro-stats}
%%%%%%%%%%%%%%%%%%%%%%%%%%%%%%%%%%%%%%%%%%%%%%%%%%%%%%%%%%%%%%%%%%%%%%%%

Our main inferential tool is a matched-pair binomial test that compares
human and LLM trajectories author by author.

\paragraph{RQ1: Drift comparisons.}
For each space $S \in \{\text{TFIDF},\text{SBERT}\}$, prompt level
$\ell \in \{\text{LV1},\text{LV2},\text{LV3}\}$, and model
$M \in \{\text{DS},\text{G4B},\text{G12B},\text{LMK}\}$, we consider all
eligible human authors $a$ and their shadow authors $a'$ generated by
model $M$ under level $\ell$.
Let $D_{a}^{(H,S)}$ and $D_{a}^{(L,S)}$ denote their
TotalDrift values in space $S$ (computed over common year pairs).
Define the indicator
\[
I_{a}^{(S,\ell,M)}
=
\mathbb{I}\!\left[
D_{a}^{(H,S)} > D_{a}^{(L,S)}
\right],
\]
and the empirical human win rate
\[
\hat{p}^{(S,\ell,M)}
=
\frac{1}{N^{(S,\ell,M)}}
\sum_{a} I_{a}^{(S,\ell,M)},
\]
where $N^{(S,\ell,M)}$ is the number of valid pairs.
We test the one-sided null
$H_{0}: p^{(S,\ell,M)} = 0.5$ versus
$H_{1}: p^{(S,\ell,M)} > 0.5$
using an exact binomial test (or its normal approximation for large $N$).
Table~\ref{tab:rq1-robustness-drift} report these win rates and
$p$-values.

Because the number of drift tests is small, we do not apply FDR correction to drift results.

\paragraph{RQ2: Cog-Emo variability comparisons (CV).}
For Cog-Emo--CV features we perform feature-wise binomial tests.
Fix a model $M$ and level $\ell$.
For each Cog-Emo feature $j \in \{1,\dots,20\}$ and each eligible author $a$,
let $c^{(H)}_{a,j}$ and $c^{(L)}_{a,j}$ denote the human and LLM CV
values for that feature. Define
\[
\begin{aligned}
I_{a,j}^{(\ell,M)}
&=
\mathbb{I}\!\left[
c^{(H)}_{a,j} > c^{(L)}_{a,j}
\right], \\
\hat{p}_{j}^{(\ell,M)}
&=
\frac{1}{N_{j}^{(\ell,M)}}
\sum_{a} I_{a,j}^{(\ell,M)}.
\end{aligned}
\]
We again test $H_{0}: p_{j}^{(\ell,M)} = 0.5$ versus
$H_{1}: p_{j}^{(\ell,M)} > 0.5$ using the binomial model.
This yields $20$ feature-wise tests per $(\ell,M)$ configuration.
The main text reports the number of features that remain significant
after FDR correction, along with mean win rates, in
Table~\ref{tab:rq2-binomial-summary}.

\paragraph{RQ2: Cog-Emo variability comparisons (RMSSD\_norm / MASD\_norm).}
For the RMSSD\_norm and MASD\_norm variants, we run the same
feature-wise matched-pair binomial tests at LV3 only.
For each metric $Z \in \{\mathrm{RMSSD\_norm},\mathrm{MASD\_norm}\}$,
model $M$, and Cog-Emo feature $j$, we define
\[
I_{a,j}^{(Z,M)}
=
\mathbb{I}\!\left[
z^{(H)}_{a,j} > z^{(L)}_{a,j}
\right],
\]
where $z^{(H)}_{a,j}$ and $z^{(L)}_{a,j}$ are the human and LLM
values of metric $Z$.
We test $H_{0}: p_{Z,M,j} = 0.5$ versus
$H_{1}: p_{Z,M,j} > 0.5$ and apply BH-FDR control within each
(metric, model) family of 20 tests.
Summary statistics (significant counts, fractions, and mean win rates)
are reported in Table~\ref{tab:cevar-rmssd-binomial-lv3} and
Table~\ref{tab:cevar-masd-binomial-lv3} in
Section~\ref{sec:cevar-ablation}.

\subsection{Geometry-based robustness checks for RQ1 (LV3)}
\label{app:geometry-binomial}

As a robustness check for RQ1 we run matched-pair binomial tests on the
three geometry descriptors defined in Appendix~\ref{app:micro-geo}
(path length, net displacement, tortuosity) across three embedding
spaces (SBERT, TF--IDF, and Cog-Emo) at LV3. These tests use the same
author-level pairing as the main total-drift analysis, but are
restricted to the three LV3 models that enter the RQ2 probes:
DeepSeek-V1 (DS), Claude~3.5 Haiku (CL35), and GPT--4o-mini (G4OM),
yielding 412 human--LLM author pairs per model. All tests are
one-sided, with $H_0: P(\text{Human}>\text{LLM}) = 0.5$ and no
multiple-comparisons correction (only nine tests per space).

\paragraph{SBERT geometry (semantic space).}
In the 384D SBERT space, humans show a strong and consistent advantage
over LLMs for both path length and net displacement across all three
models. Human win rates for SBERT path length range from $0.748$ to
$0.852$ (DS: $0.830$, CL35: $0.852$, G4OM: $0.748$), and for SBERT net
displacement from $0.697$ to $0.745$, with all six tests significant at
$\alpha = 0.05$ (all $p < 10^{-15}$). This confirms that human
trajectories not only move more year-to-year in semantic space but also
end up farther from their starting points than their matched LLM
trajectories.

SBERT tortuosity exhibits a much weaker pattern: human win rates cluster
around $0.49$--$0.54$, with only a single model (DS) showing a marginal
Human~$>$~LLM effect (win rate $0.544$, $p = 0.042$). Thus the main
semantic difference lies in how far trajectories move, not in the
curviness of their paths.

\paragraph{TF--IDF geometry (lexical space).}
In the 10D TF--IDF space, the direction of the effect reverses. For both
path length and net displacement, LLM trajectories typically exceed
human trajectories, with human win rates in the $0.19$--$0.31$ range
(path length: $0.194$--$0.308$; net displacement: $0.226$--$0.311$
across the three models). Under our one-sided Human~$>$~LLM tests these
comparisons are not significant (all $p = 1.0$), but the direction is
stable and contrasts sharply with SBERT: LLMs exhibit more cumulative
movement and larger end--point separation in lexical space.

TF--IDF tortuosity is close to symmetric: human win rates fall between
$0.500$ and $0.536$ with no significant Human~$>$~LLM effects
($p > 0.05$ for all three models). This suggests that the relative
curvature of lexical trajectories is broadly similar for humans and
LLMs, even though LLMs move farther lexically overall.

\paragraph{Cog-Emo geometry (cognitive--emotional space).}
In the 20D Cog-Emo space (Big Five--NELA features without z-score
normalization), humans again show a strong advantage for both path
length and net displacement. For Cog-Emo path length, human win rates span
$0.757$ (DS), $0.825$ (CL35), and $0.973$ (G4OM), with all three tests
highly significant ($p \le 10^{-27}$). For Cog-Emo net displacement, human
win rates are $0.633$ (DS), $0.743$ (CL35), and $0.947$ (G4OM), again
all significant with $p \le 10^{-8}$. These results mirror the SBERT
pattern and show that human trajectories traverse substantially larger
distances, and end farther from where they started, in
cognitive--emotional space.

Cog-Emo tortuosity, like its SBERT and TF--IDF counterparts, exhibits only
weak and mixed behavior: human win rates are close to $0.5$
($0.432$--$0.524$ across models), and none of the Human~$>$~LLM tests is
significant. Path curvature thus appears comparable across humans and
LLMs, even though the scale of movement differs sharply.

\paragraph{Semantic--lexical--cognitive dissociation.}
Taken together, these geometry-based tests reinforce and sharpen the
main RQ1 finding. Human trajectories exhibit substantially larger
evolution in \emph{semantic} (SBERT) and \emph{cognitive--emotional}
(Cog-Emo) spaces, whereas LLM trajectories tend to move more in
\emph{lexical} (TF--IDF) space. Tortuosity shows no consistent
Human~$>$~LLM advantage in any space, indicating that the key
differences lie in how far trajectories travel rather than in how
curved their paths are. The additional geometry descriptors therefore
support a three-way dissociation: humans change more in meaning and
cognitive--emotional profile over time, whereas LLMs compensate by
varying word choice while keeping their underlying semantic and
cognitive--emotional trajectories comparatively flattened.
\subsection{Drift robustness across prompt levels and encoders}
\label{app:drift-robustness}

Beyond the LV3 results reported in Section~\ref{sec:results}, we verify
that the Human $>$ LLM drift pattern in semantic and cognitive--emotional
spaces is robust to both prompt design and embedding choice.

\paragraph{Prompt-level robustness (LV1--LV3).}
For Gemma3 4B, Gemma3 12B, and Llama 4 (LMK), we recompute
TotalDrift-based matched-pair binomial tests at all three prompt levels
(LV1, LV2, LV3) in SBERT and TF--IDF spaces. Table~\ref{tab:rq1-robustness-drift}
summarizes mean Human win rates across levels.

\begin{table}[t]
\centering
\small
\caption{RQ1 drift robustness across prompt levels.
Mean Human win rates $\hat{p}$ by space and level (averaged over
Gemma3 4B, Gemma3 12B, DeepSeek and Llama 4).}
\label{tab:rq1-robustness-drift}
\begin{tabular}{lccc}
\toprule
\textbf{Space} & \textbf{LV1} & \textbf{LV2} & \textbf{LV3} \\
\midrule
SBERT & 0.78 & 0.78 & 0.80 \\
TF--IDF & 0.26 & 0.22 & 0.26 \\
Cog-Emo (drift) & 0.71 & 0.68 & 0.79 \\
\bottomrule
\end{tabular}
\end{table}

Across all three levels, SBERT and Cog-Emo drift show a stable
Human~$>$~LLM pattern (mean win rates $\approx 0.68$--$0.80$), whereas
TF--IDF exhibits the opposite direction (mean win rates
$0.22$--$0.26$, i.e., LLM $>$ Human). Importantly, changing from LV1 to
LV3 does not alter the qualitative ordering: humans move more in
semantic and Cog-Emo spaces, while LLMs move more in lexical space.

\paragraph{Encoder robustness (E5-Large vs.\ MiniLM SBERT).}
To test sensitivity to the choice of semantic encoder, we recompute LV3
drift in SBERT space using E5-Large (1024d) instead of
all-MiniLM-L6-v2 (384d). For E5-Large, Human win rates remain
substantially above chance for all six LV3 models (mean
$\hat{p} \approx 0.68$), and every Human~$>$~LLM binomial test is
significant at $\alpha=0.05$. While win rates are slightly lower than
for MiniLM, the direction and strength of the effect are preserved,
indicating that the Human $>$ LLM semantic-drift pattern is not an
artifact of a particular SBERT architecture.
\subsection{Group-level cross-validation and leakage prevention.}
All predictive probes use GroupKFold with the \texttt{author\_id} as the
group identifier. For each author $a$, the human trajectory ($T_a^{H}$)
and its matched LLM trajectories ($T_{a}^{L,1},\dots,T_{a}^{L,4}$) share
the exact same \texttt{author\_id}. Therefore, the grouping constraint
forces all samples associated with author $a$—both human and LLM—into the
same fold.

This guarantees that:
(i) no fold ever contains human samples of author $a$ in training while
containing that author's LLM samples in testing (or vice versa);
(ii) the classifier never receives partial information about an author's
trajectory during training; and
(iii) all evaluations reflect generalization across authors rather than
within-author memorization.

This design eliminates any form of cross-fold data leakage and ensures
that predictive performance is attributable to temporal-dynamic
differences rather than identity-level artifacts.

\section{Multiple Comparisons and Effect Sizes}
\label{app:stats-multiple}

This appendix details the multiple-testing correction and effect-size
reporting used for Cog-Emo variability feature-wise binomial tests in
Section~\ref{sec:results} and Section~\ref{sec:cevar-ablation}.

\subsection{Families of Hypotheses}

For Cog-Emo--CV we consider, for each prompt level
$\ell \in \{\mathrm{LV1},\mathrm{LV2},\mathrm{LV3}\}$ and each model
\[
M \in 
\begin{aligned}
\{&\text{DeepSeek-V1},\ \text{Gemma3 4B},\ \text{Gemma3 12B},\\
  &\text{Llama 4},\ \text{Claude~3.5 Haiku},\ \text{GPT--4o-mini}\},
\end{aligned}
\]
the family of 20 hypotheses:
\[
\begin{aligned}
H_{0}^{(\ell,M,j)} &: \; p_{\ell,M,j} = 0.5, \\
H_{1}^{(\ell,M,j)} &: \; p_{\ell,M,j} > 0.5.
\end{aligned}
\]
where $p_{\ell,M,j}$ denotes the probability that the human CV exceeds
the matched LLM CV for feature $j$ under configuration $(\ell,M)$.

For RMSSD\_norm and MASD\_norm we analogously define, for each metric
$Z \in \{\mathrm{RMSSD\_norm},\mathrm{MASD\_norm}\}$ and model $M$ at
LV3, a family of 20 hypotheses
$H_{0}^{(Z,M,j)}: p_{Z,M,j} = 0.5$ vs.\
$H_{1}^{(Z,M,j)}: p_{Z,M,j} > 0.5$.

\subsection{Benjamini--Hochberg FDR Control}

We control the false discovery rate (FDR) at $q=0.05$ using the
Benjamini--Hochberg (BH) procedure separately within each family of
20 tests:

- for Cog-Emo--CV: within each $(\ell,M)$ configuration;
- for RMSSD\_norm and MASD\_norm: within each $(Z,M)$ configuration.

Let $p_{(1)} \le \cdots \le p_{(20)}$ denote the
sorted raw $p$-values in a given family.
BH declares feature $j$ significant if
\[
p_{(i)} \le \frac{i}{20} q
\]
for the largest such index $i$.
Counts of significant features (``Sig.\ (20)'') and their fractions
for Cog-Emo--CV are reported in
Table~\ref{tab:rq2-binomial-summary}; the corresponding RMSSD\_norm
and MASD\_norm summaries appear in
Table~\ref{tab:cevar-rmssd-binomial-lv3} and
Table~\ref{tab:cevar-masd-binomial-lv3}.

\paragraph{Note.}
Drift-based tests (RQ1) involve only a small fixed number of hypotheses
per space and level, so we report uncorrected $p$-values for those
comparisons.

\subsection{Effect Size for Win-Rate Deviations (Cohen's $h$)}

To quantify the magnitude of deviation of the observed win rate
$\hat{p}$ from chance ($0.5$) independent of sample size, we
compute Cohen's $h$:
\[
h
=
2\left(\arcsin\sqrt{\hat{p}} - \arcsin\sqrt{0.5}\right).
\]
Positive $h$ indicates a human advantage
($\hat{p} > 0.5$), whereas negative $h$ indicates an LLM advantage.
When summarizing effect sizes across groups of features (e.g., cognitive
vs.\ stylistic vs.\ emotional), we report mean $h$ within each group.
Standard qualitative thresholds ($|h|\approx 0.2/0.5/0.8$ for
small/medium/large) are used only as descriptive guidance.

\subsection{Full Figure for 20 Cog-Emo-CV Difference }
\label{app:cecv-differnece-figures}
\begin{figure*}[t]
     \centering
     \includegraphics[width=\textwidth]{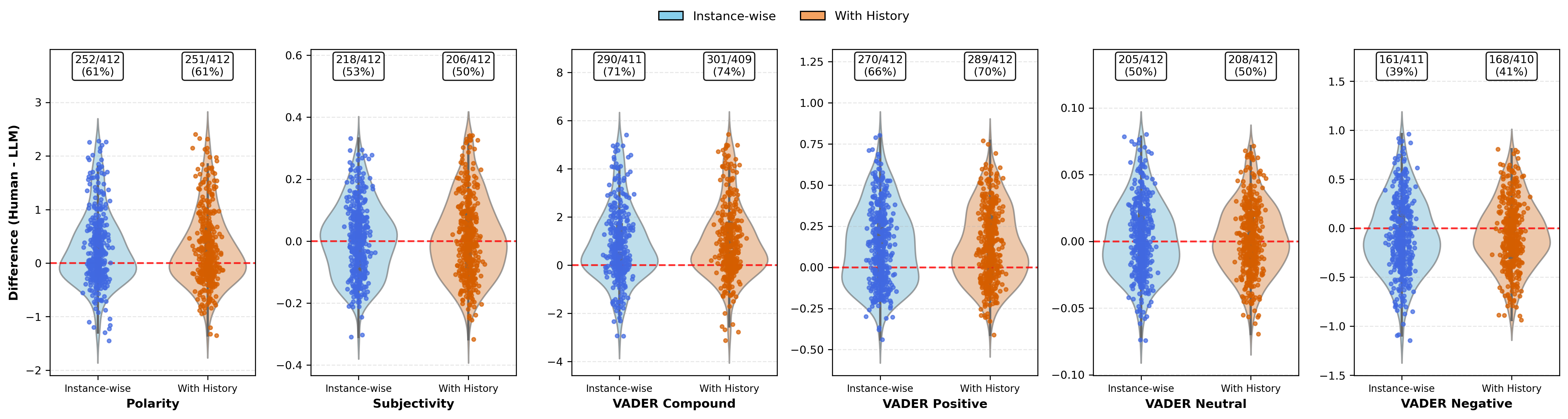}
     \caption{Cog-Emo CV differences for sentiment features 
     (polarity, subjectivity, VADER scores), comparing instance-wise and history-augmented conditions}. Positive values: Human $>$ LLM.
     \label{fig:cv-emotional}
 \end{figure*}

\begin{figure*}[t]
     \centering
    \includegraphics[width=\textwidth]{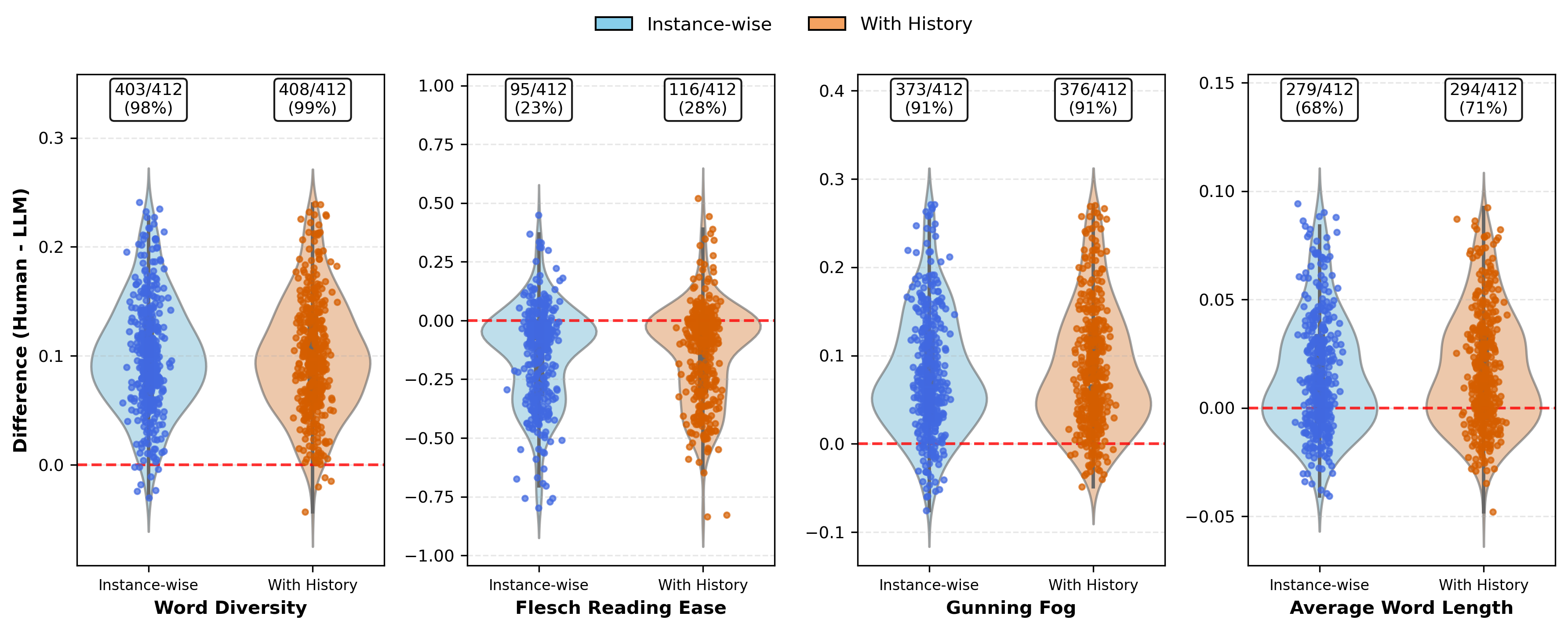}
     \caption{Cog-Emo CV differences for stylistic features, Group 1 
     (lexical diversity, readability, length, POS ratios), comparing instance-wise and history-augmented conditions}. Positive values: Human $>$ LLM.
     \label{fig:cv-stylistic-1}
 \end{figure*}

\begin{figure*}[t]
     \centering
     \includegraphics[width=\textwidth]{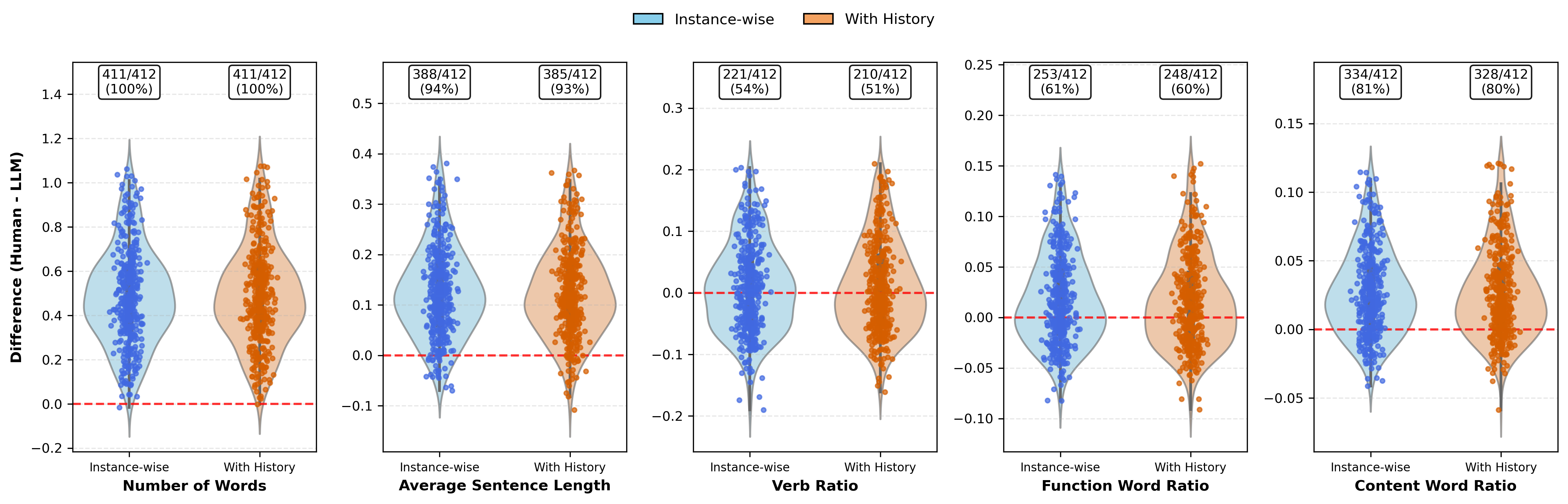}
     \caption{Cog-Emo CV differences for stylistic features, Group 2, comparing instance-wise and history-augmented conditions}. 
     Positive values: Human $>$ LLM.
     \label{fig:cv-stylistic-2}
 \end{figure*}

Figures \ref{fig:cv-emotional}, \ref{fig:cv-stylistic-1}, and \ref{fig:cv-stylistic-2} report the complete Human–LLM difference based on GPT-4o-mini, comparing instance-wise and history-augmented conditions.

\section{Cog-Emo--CV Classifier: Class Balance, Per-Class Metrics}
\label{app:cecvc}
%%%%%%%%%%%%%%%%%%%%%%%%%%%%%%%%%%%%%%%%%%%%%%%%%%%%%%%%%%%%%%%%%%%%%%%%

This appendix reports additional diagnostic information for the
Cog-Emo--CV classifier used in the predictive probes in
Section~\ref{sec:results}. All analyses are conducted under
LV3 using a Random Forest classifier with 5-fold GroupKFold
(grouped by \texttt{author\_id}), ensuring that all human and LLM
shadow trajectories of the same author remain within the same fold.

\paragraph{Class balance.}
Because each human author is matched with three LLM shadow authors
(DeepSeek-V1, Claude~3.5 Haiku, and GPT--4o-mini),
the classification dataset contains a fixed 1:3 ratio:
\[
\text{Human} = 412 \ (\approx 25\%), \qquad
\text{LLM} = 1{,}236 \ (\approx 75\%),
\]
for a total of 1{,}648 samples.
This proportion is constant across the academic, blogs, and news
domains. GroupKFold prevents any cross-fold leakage of author identity.

\paragraph{Overall and per-class metrics.}
Averaged over the five cross-validation folds, the classifier achieves:
\[
\begin{aligned}
\text{Accuracy}   &= 0.9363 \pm 0.0117,\\
\text{ROC--AUC}   &= 0.9770 \pm 0.0085,\\
\text{F1 (macro)} &= 0.8631 \pm 0.0283.
\end{aligned}
\]
For the minority (human) class, we obtain:
\[
\begin{aligned}
\text{Recall}_{\text{Human}} = 0.8082 \pm 0.0473.
\end{aligned}
\]
LLM precision and recall remain high (both above 0.9) with near-perfect
LLM recall, reflecting that the classifier rarely confuses LLM trajectories
for human ones.

\paragraph{Interpretation.}
The combination of high accuracy, ROC--AUC, and macro-F1, together with
human recall around 0.81, indicates that human and LLM trajectories are
well separated in Cog-Emo--CV space. The model operates in a regime
where both precision and recall for the human class are reasonably high,
rather than relying on a degenerate threshold or extreme conservatism.
\paragraph{Summary.}
These diagnostics confirm that the Cog-Emo--CV classifier is not affected by
class leakage or pathological imbalance: the observed performance reflects
genuine temporal-dynamic differences between human and LLM trajectories,
rather than identity-level artifacts or overfitting to particular authors.
\subsection{Domain-wise CE--CV Classification (LV3)}
\label{app:cecv-domain}

To ensure the pooled CE--CV probe is not dominated by a single domain,
we repeat the LV3 classification experiment separately on Academic, Blogs,
and News. We keep the same configuration as in the pooled probe:
Random Forest, 20 CE--CV features, and 5-fold GroupKFold by \texttt{author\_id}.
Table~\ref{tab:cecv-domain} reports mean performance across folds.

\begin{table}[t]
\centering
\small
\caption{Domain-wise CE--CV classification performance (LV3; pooled DS/CL35/G4OM within each domain).}
\label{tab:cecv-domain}
\setlength{\tabcolsep}{4pt}
\begin{tabular}{lcccc}
\toprule
\textbf{Domain} & \textbf{Acc.} & \textbf{AUC} & \textbf{F1} & \textbf{Rec.(H)} \\
\midrule
Academic & 0.9300 & 0.9613 & 0.8470 & 0.7700 \\
Blogs    & 0.9641 & 0.9960 & 0.9252 & 0.8923 \\
News     & 1.0000 & 1.0000 & 1.0000 & 1.0000 \\
\bottomrule
\end{tabular}
\end{table}

Across all three domains, CE--CV features remain highly discriminative and
the direction of effect is consistent with the pooled result, indicating that
the pooled performance is not driven by a single domain.

\subsection{Domain-wise Ablation: Removing Length/Readability CV Features}
\label{app:cecv-domain-ablation}

We further repeat the domain-wise probes after removing the same six
length/readability CV features used in Appendix~\ref{app:cecv-length-ablation}.
Table~\ref{tab:cecv-domain-ablation} summarizes performance.

\begin{table}[t]
\centering
\small
\caption{Domain-wise CE--CV classification after removing six length/readability features (LV3).}
\label{tab:cecv-domain-ablation}
\setlength{\tabcolsep}{4pt}
\begin{tabular}{lcccc}
\toprule
\textbf{Domain} & \textbf{Acc.} & \textbf{AUC} & \textbf{F1} & \textbf{Rec.(H)} \\
\midrule
Academic & 0.8625 & 0.9120 & 0.6797 & 0.5900 \\
Blogs    & 0.9410 & 0.9853 & 0.8706 & 0.8103 \\
News     & 0.9705 & 0.9939 & 0.9355 & 0.8906 \\
\bottomrule
\end{tabular}
\end{table}

Although performance decreases as expected, all domains remain well above chance,
confirming that the human--LLM separation is not explained solely by surface-form
(length/readability) variability.

\section{Cog-Emo-VAR Validity and Robustness Analyses}
\label{sec:cevar-ablation}
%%%%%%%%%%%%%%%%%%%%%%%%%%%%%%%%%%%%%%%%%%%%%%%%%%%%%%%%%%%%%%%%%%%%%%%%

This section summarizes additional validity and robustness analyses for
the Cog-Emo variability operator (Cog-Emo-VAR) beyond the main CV-based
results in Section~\ref{sec:results}. We evaluate whether two alternative
temporal-variability descriptors—\texttt{RMSSD\_norm} and
\texttt{MASD\_norm}—preserve the Human $>$ LLM pattern observed under
the CV operator and whether they maintain predictive utility in a
machine-learning probe.

\subsection{Ablation of Length and Stylistic Features}
\label{app:cecv-length-ablation}

To verify that the Cog-Emo--CV classifier is not driven purely by
length- and structure-related signals, we perform an ablation study in
which six length/stylistic CV features are removed:
\texttt{num\_words\_cv},
\texttt{avg\_sentence\_length\_cv},
\texttt{flesch\_reading\_ease\_cv},
\texttt{gunning\_fog\_cv},
\texttt{word\_diversity\_cv}, and
\texttt{average\_word\_length\_cv}.
The remaining 14 features consist of 5 personality proxies, 6
affective/VADER features, and 3 POS-ratio features
(\texttt{verb\_ratio\_cv}, \texttt{function\_word\_ratio\_cv},
\texttt{content\_word\_ratio\_cv}).

We reuse the LV3 pooled setting (DeepSeek-V1, Claude~3.5 Haiku,
GPT--4o-mini; 412 Human vs.\ 1,236 LLM trajectories; $N{=}1{,}648$) and
the same Random Forest configuration as in Section~\ref{sec:results}
(300 trees, class-balanced weights, 5-fold GroupKFold by
\texttt{author\_id}). Table~\ref{tab:cecv-length-ablation} compares the
20-feature and 14-feature models.

\begin{table}[t]
\centering
\small
\caption{CE--CV classifier performance with the full 20-feature set vs.\
an ablated 14-feature set that removes length/readability features
(pooled DS/CL35/G4OM setting).}
\label{tab:cecv-length-ablation}
\setlength{\tabcolsep}{4pt}
\begin{tabular}{lcccc}
\toprule
\textbf{Feature set} & \textbf{Acc.} & \textbf{AUC} & \textbf{F1} & \textbf{Rec.(H)} \\
\midrule
20 features (full)   & 0.9363 & 0.9770 & 0.8631 & 0.8082 \\
14 features (ablated)& 0.9065 & 0.9467 & 0.7905 & 0.7063 \\
\bottomrule
\end{tabular}
\end{table}

Removing the six length/readability features yields a moderate but
consistent performance drop (Accuracy $-3.2$ points, AUC $-3.1$ points,
F1 $-8.4$ points, Human recall $-12.6$ points), yet the ablated model
still achieves $>0.90$ accuracy and $>0.94$ AUC. Feature-importance
analysis shows that, in the ablated model, personality proxies become
dominant (accounting for $\sim 52\%$ of total importance), followed by
VADER sentiment components ($\sim 24\%$) and POS-ratio features
($\sim 16\%$), with Agreeableness and Neuroticism emerging as the two
strongest individual predictors.

Taken together, these results indicate that length and readability
variability provide complementary signal, but are not solely responsible
for the classifier's performance. Even when all length- and
structure-related CV features are removed, human and LLM trajectories
remain highly separable based on cognitive--emotional and POS-ratio
variability alone.

\subsection{RMSSD\_norm and MASD\_norm Binomial Tests (LV3)}

We run matched-pair binomial tests on the 20 Cog-Emo features using
RMSSD\_norm and MASD\_norm at LV3, across the same 6 LLMs (DeepSeek-V1, Gemma3 4B, Gemma3 12B, Llama 4, Claude~3.5, GPT--4o-mini) and 412 authors. As in the Cog-Emo--CV setting, we test
$H_0: P(\text{Human} > \text{LLM}) = 0.5$ versus
$H_1: P(\text{Human} > \text{LLM}) > 0.5$ and apply BH-FDR correction
within each (metric, model) family.

Table~\ref{tab:cevar-rmssd-binomial-lv3} and
Table~\ref{tab:cevar-masd-binomial-lv3} summarize the number of
significant features and mean win rates.

\begin{table}[t]
\centering
\small
\caption{RMSSD\_norm (LV3). Summary of feature-wise binomial tests over
20 Cog-Emo features per model (FDR $q<0.05$).}
\label{tab:cevar-rmssd-binomial-lv3}
\begin{tabular}{lccc}
\toprule
\textbf{Model} & \textbf{Sig.\ (20)} & \textbf{Fraction} & \textbf{Mean Win Rate} \\
\midrule
DeepSeek-V1      & 16 & 0.80 & 0.635 \\
Gemma3 4B     & 13 & 0.65 & 0.668 \\
Gemma3 12B    & 17 & 0.85 & 0.653 \\
Llama 4       &  9 & 0.45 & 0.543 \\
Claude 3.5    & 16 & 0.80 & 0.660 \\
GPT--4o-mini  & 14 & 0.70 & 0.690 \\
\midrule
\textbf{Overall} & 85 & 0.71 & -- \\
\bottomrule
\end{tabular}
\end{table}

\begin{table}[t]
\centering
\small
\caption{MASD\_norm (LV3). Summary of feature-wise binomial tests over
20 Cog-Emo features per model (FDR $q<0.05$).}
\label{tab:cevar-masd-binomial-lv3}
\begin{tabular}{lccc}
\toprule
\textbf{Model} & \textbf{Sig.\ (20)} & \textbf{Fraction} & \textbf{Mean Win Rate} \\
\midrule
DeepSeek-V1     & 14 & 0.70 & 0.619 \\
Gemma3 4B     & 13 & 0.65 & 0.659 \\
Gemma3 12B    & 16 & 0.80 & 0.633 \\
Llama 4       &  9 & 0.45 & 0.530 \\
Claude 3.5    & 16 & 0.80 & 0.646 \\
GPT--4o-mini  & 14 & 0.70 & 0.683 \\
\midrule
\textbf{Overall} & 82 & 0.68 & -- \\
\bottomrule
\end{tabular}
\end{table}

Across both metrics, the pattern closely mirrors Cog-Emo--CV:

\begin{itemize}
    \item \textbf{Strong overall signal.}  
    RMSSD\_norm yields 85/120 (70.8\%) significant tests, and
    MASD\_norm yields 82/120 (68.3\%) significant tests after FDR
    correction, confirming a robust Human $>$ LLM advantage in temporal
    variability.

    \item \textbf{Consistent model ranking.}  
    Gemma3 12B shows the strongest signal (80--85\% of features significant),
    DeepSeek-V1, Gemma3 4B, Claude~3.5, and GPT--4o-mini cluster in the
    65--80\% range, and Llama~4 consistently exhibits the weakest but still
    generally positive pattern (45\% of features).

    \item \textbf{Stable feature-level structure.}  
    Several features are consistently significant across six
    models and both metrics, including Neuroticism, Openness,
    average sentence length, and Gunning Fog. These echo the high-importance
    dimensions in the Cog-Emo--CV classifier
    (Table~\ref{tab:rq2-importance}).

    \item \textbf{Features with reversed or weak signal.}  
    Flesch Reading Ease consistently shows Human $<$ LLM
    (win rates around 0.11--0.35) across models, and
    \texttt{vader\_compound} exhibits mixed behavior (Human $>$ LLM for
    some models, LLM $>$ Human for others). These exceptions highlight
    that not all affective/readability dimensions carry reliable
    human-favoring temporal signal.
\end{itemize}

Overall, RMSSD\_norm and MASD\_norm corroborate the main Cog-Emo--CV finding:
\emph{human trajectories exhibit substantially higher temporal
variability than LLM trajectories across a broad set of
cognitive--emotional features}, and this pattern is robust to the choice
of variability operator.

\subsection{Predictive Power of RMSSD and MASD}
\label{app:cevar-ml}

To further validate the robustness of our findings, we evaluated whether these alternative variability descriptors (\texttt{RMSSD\_norm} and \texttt{MASD\_norm}) possess sufficient discriminative power to distinguish human from LLM trajectories in a predictive setting. We replicated the machine learning probe described in Section~\ref{sec:results} using the exact same configuration (Random Forest, $N=300$, balanced class weights, 5-fold GroupKFold) at LV3.

Table~\ref{tab:cevar-ml-comparison} reports the classification metrics.
The CV-based operator achieves strong performance (Accuracy $0.9363$,
AUC $0.9770$, F1 $0.8631$, Human Recall $0.8082$) on the combined
DeepSeek-V1/Claude 3.5 Haiku/GPT-4o-mini LV3 setting. Both \texttt{RMSSD\_norm} and
\texttt{MASD\_norm} also achieve high accuracy ($>0.90$) and AUC
($>0.94$). Although their scores are modestly lower than those of the
CV-based probe, the overall performance indicates that the
temporal-flattening signal remains detectable across substantially
different variability operators.

\begin{table}[h]
\centering
\small
\caption{Classification performance comparison (LV3) using different temporal variability operators.}
\label{tab:cevar-ml-comparison}

% --- reduce column padding ---
\setlength{\tabcolsep}{3.5pt}

\begin{tabular}{lcccc}
\toprule
\textbf{Metric} & \textbf{Acc.} & \textbf{AUC} & \textbf{F1} & \textbf{Rec.(H)} \\
\midrule
\textbf{Cog-Emo--CV (Main)} & \textbf{0.936} & \textbf{0.977} & \textbf{0.863} & \textbf{0.808} \\
RMSSD\_norm            & 0.913          & 0.968          & 0.811          & 0.755 \\
MASD\_norm             & 0.911          & 0.962          & 0.806          & 0.735 \\
\bottomrule
\end{tabular}
\end{table}

\paragraph{Feature Importance.}
Across both RMSSD\_norm and MASD\_norm classifiers, 
\texttt{avg\_sentence\_length} consistently emerges as the single most 
predictive feature, replicating the CE--CV results and underscoring the 
central role of stylistic variability in distinguishing human trajectories. 
Personality-linked features---especially \texttt{Agreeableness} and 
\texttt{Neuroticism}---also remain among the high-importance dimensions, 
though their relative rankings differ across operators: 
\texttt{Agreeableness} is the second most important feature under 
RMSSD\_norm, whereas MASD\_norm elevates \texttt{Neuroticism} and 
lexical-diversity features (e.g., \texttt{word\_diversity}, 
\texttt{num\_words}). 
This convergence across operators further reinforces that both stylistic and 
cognitive–emotional variability contribute robustly to identifying authentic 
human temporal dynamics.

\subsection{Additional CE--CV classifier robustness tests}
\label{app:cecv-robustness}

Finally, we verified that the CE--CV classifier is robust to changes in
prompt level and the underlying LLM pool. All robustness probes reuse
the 20-dimensional Cog-Emo--CV feature vector and the same
Random Forest classifier with 5-fold GroupKFold (grouped by
\texttt{author\_id}).

\paragraph{Prompt-level robustness (LV1--LV3, DS/G4B/G12B/LMK).}
We first recompute the CE--CV probe on an independent four-model pool
(DeepSeek-V1, Gemma3 4B, Gemma3 12B, Llama 4) at LV1--LV3, pooling
academic, blogs, and news domains. Table~\ref{tab:cecv-lv-robustness}
summarizes accuracy, ROC--AUC, macro-F1, and human recall averaged
over the four models at each level.

\begin{table}[t]
\centering
\small
\caption{CE--CV classifier robustness across prompt levels (LV1--LV3). 
Mean performance over 5 GroupKFold splits.}
\label{tab:cecv-lv-robustness}
\setlength{\tabcolsep}{3pt}
\begin{tabular}{lcccc}
\toprule
\textbf{Level} & \textbf{Acc.} & \textbf{AUC} & \textbf{F1} & \textbf{Rec.} \\
\midrule
LV1 (Zero-shot)   & 0.903 & 0.956 & 0.705 & 0.580 \\
LV2 (Persona)     & 0.910 & 0.954 & 0.724 & 0.592 \\
LV3 (Pers.+Ex.)   & 0.921 & 0.963 & 0.763 & 0.636 \\
\bottomrule
\end{tabular}
\end{table}

Performance remains strong at all three levels (Accuracy $>0.90$,
AUC $>0.95$), with F1 and human recall improving monotonically from LV1
to LV3. Thus, the temporal-dynamic signal exploited by CE--CV does not
depend on a specific prompt level.

\subsection{CV Robustness in TF--IDF and SBERT Spaces}
\label{app:variance-tfidf-sbert}

To verify that the Human $>$ LLM temporal-variability pattern is not tied to
the Cog-Emo feature space, we computed \emph{coefficient-of-variation (CV)}
based temporal descriptors directly in TF--IDF (10D) and SBERT (384D)
embedding spaces at LV3. For each author and model, we obtained yearly
centroids and computed per-dimension CV over years 
$\mathrm{CV} = \sigma / |\mu|$. We then performed matched-pair binomial tests
for each embedding dimension comparing Human vs.\ LLM CV across 412 author
pairs (one-sided Human $>$ LLM). Significant dimensions were determined using
BH-FDR ($q < 0.05$).

\begin{table}[t]
\centering
\small
\caption{CV-based Human $>$ LLM binomial tests in TF--IDF and SBERT spaces
(10 and 384 dimensions respectively). Reported are the fraction of
dimensions where Human CV significantly exceeds LLM CV (BH-FDR $q<0.05$)
and mean Human win rate across dimensions.}
\label{tab:variance-tfidf-sbert}
\setlength{\tabcolsep}{5pt}
\begin{tabular}{lcccc}
\toprule
\textbf{Model} & \textbf{Space} & \textbf{Dims} & \textbf{Sig.} & \textbf{WinRate} \\
\midrule
DeepSeek-V1   & TF--IDF & 10  & 0.20 & 0.47 \\
Claude~3.5    & TF--IDF & 10  & 0.10 & 0.43 \\
GPT--4o-mini  & TF--IDF & 10  & 0.30 & 0.49 \\
\midrule
DeepSeek-V1   & SBERT   & 384 & 0.57 & 0.57 \\
Claude~3.5    & SBERT   & 384 & 0.65 & 0.58 \\
GPT--4o-mini  & SBERT   & 384 & 0.64 & 0.58 \\
\bottomrule
\end{tabular}
\end{table}

\paragraph{Summary.}
Across LV3, TF--IDF-based CV shows only weak or mixed Human advantages:
mean Human win rates remain close to chance (0.43–0.49) with only 10–30\%
of dimensions significant after FDR correction. In contrast, SBERT-based CV
reveals a strong and consistent Human $>$ LLM pattern across all three
models: 57–65\% of the 384 dimensions show significantly greater Human
temporal CV, and mean Human win rates remain around 0.57–0.58. These
results demonstrate that temporal flattening is not tied to Cog-Emo
features: the Human $>$ LLM variability pattern persists in high-dimensional
semantic spaces, reinforcing its robustness.

\end{document}